\documentclass{antgroup}
\PassOptionsToPackage{numbers,sort&compress}{natbib}
\usepackage{antgroup}
\setcitestyle{numbers,square,comma}

\usepackage{amsmath,amssymb,mathtools,bm}
\usepackage{xspace}
\usepackage{enumitem}
\usepackage{booktabs}
\usepackage{graphicx}
\usepackage{caption}
\usepackage{cleveref}
\usepackage{needspace}

\pretocmd{\section}{\Needspace{5\baselineskip}}{}{}
\pretocmd{\subsection}{\Needspace{3\baselineskip}}{}{}
\setlist[itemize]{leftmargin=1.5em,itemsep=2pt,topsep=4pt}
\newcommand{\cwm}{Invoker-Render\xspace}

\title{Code World Model 2.0:\\
Towards Infinite Controllable Video Generation}
\author{%
\textbf{Ruoyu Wang}$^{1,2,\ast}$\quad
\textbf{Hao Zhong}$^{1,3,\ast}$\quad
\textbf{Yanming Yang}$^{1,2,\ast}$\quad
\textbf{Jiahao Tian}$^{1,2,\ast}$\quad
\textbf{Lin Liu}$^{1,\ast,\dagger}$\\[3pt]
\textbf{Biao Gong}$^{1}$\quad
\textbf{Guoshan Lu}$^{1}$\quad
\textbf{Yiding Tian}$^{1}$\quad
\textbf{Chunhua Shen}$^{3,\ddagger}$\quad
\textbf{Zhenzhong Lan}$^{1,2,\ddagger}$\quad
\textbf{Chi Zhang}$^{2,\ddagger}$%
}
\affiliation{\selectfont
$^{1}$Inclusion AI, Ant Group\quad
$^{2}$Westlake University\quad
$^{3}$Zhejiang University\\[2pt]
{\footnotesize $^{\ast}$Equal contribution\quad $^{\dagger}$Project leader\quad $^{\ddagger}$Corresponding author}%
}
\hypersetup{pdftitle={Code World Model 2.0: Towards Infinite Controllable Video Generation}}

\begin{document}
\maketitle
\vspace{-4.5mm}
{\centering\small
\href{https://www.inclusion-ai.org/CodeWorldModel/}{\textbf{Project Page}}
\hspace{0.9em}\textcolor{gray}{\textbullet}\hspace{0.9em}
\href{https://github.com/inclusionAI/CodeWorldModel}{\textbf{GitHub}}
\hspace{0.9em}\textcolor{gray}{\textbullet}\hspace{0.9em}
\href{https://huggingface.co/inclusionAI/invoker-render-1.0}{\textbf{Hugging Face}}\par}
\vspace{2.5mm}

\begin{abstract}
Visual proxies decouple explicit world-state simulation from photorealistic video synthesis, offering a controllable interface for both offline clip creation and real-time interactive streaming. Yet a practical renderer must address two challenges. Upstream proxies vary from coarse layouts to fine articulated geometry depending on the interaction, and streaming inference must operate in a few causal steps without accumulating drift across self-generated history. We present \textbf{\cwm}, pairing a bidirectional creation model with a causal four-step streaming model under a shared conditioning interface over structural proxies, an appearance reference, and language. Trained on an aligned $\sim$200-hour multi-granularity corpus, \cwm honors fine geometric constraints when supplied while synthesizing plausible local motion under coarse proxies, enabling runtime selection of input detail. To bridge bidirectional creation and real-time streaming without compounding step compression and rollout drift, we introduce \textbf{Progressive Forcing (ProForcing)}. ProForcing first trains a causal predictor and distills it into a four-step renderer using histories derived from training videos. Training then shifts to continuous self-generated rollouts. A pipelined serving stack supports efficient streaming inference. Experiments show versatile control across proxy families, granularities, and camera paths, consistent gains over recent streaming baselines, and stable 30-minute racing rollouts that preserve scene and subject appearance.
\end{abstract}

\par\noindent\begin{minipage}{\linewidth}
  \centering
  \includegraphics[width=\linewidth,height=192pt,keepaspectratio]{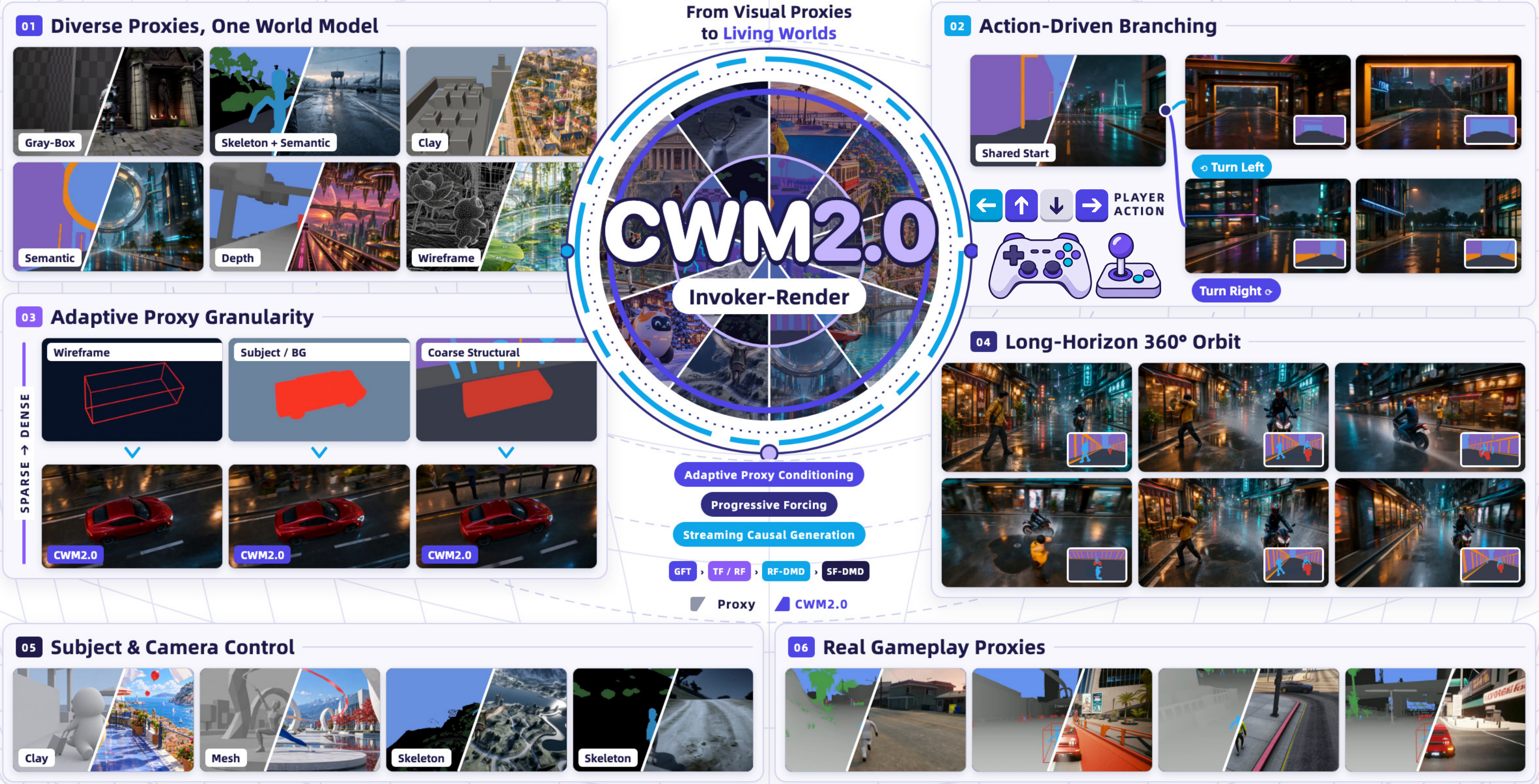}
  \captionsetup{font=small,skip=4pt,hypcap=false}
  \captionof{figure}{\textbf{Controllable video rendering with \cwm.} Conditioned on multi-granularity structural proxies, an appearance reference, and text, our framework unifies bidirectional video creation and four-step causal streaming.}
  \label{fig:teaser}
\end{minipage}\par

\newpage

\section{Introduction}
\label{sec:introduction}

Interactive video world models must synthesize coherent observations as user actions alter the environment \citep{bruce2024genie,valevski2024gamengen,zhang2025matrixgame,sun2026worldplay}. Rather than simulating both dynamics and appearance in one network, an emerging paradigm maintains explicit state in an upstream engine and delegates visual synthesis to a generative renderer \citep{chen2026cwm,huang2026pwm,wu2026boxworld}. Visual and geometric proxies bridge the two stages by prescribing camera paths, layout, and entity motion while leaving photorealistic appearance to the video model \citep{liang2026codevideomodel,meng2026marionette,chen2026dar,liang2025diffusionrenderer}. Recent systems pair such structural cues with appearance references for offline authoring and real-time rendering \citep{fang2026solarrenderer,lu2026streamrenderh3,liu2026anygamerenderer,huang2026edenlive,zhan2026magpie}.

In practice, the structural detail appropriate for conditioning depends on the interaction. Broad navigation often requires only coarse scene geometry and root trajectories, whereas close-range interactions demand articulated poses or finer surface structure. Detailed proxies supply tighter constraints at higher construction and transmission cost, while coarse proxies are lightweight to produce yet leave local geometry and motion under-specified. A capable renderer must interpret both regimes faithfully. For example, a rigid mannequin that tracks a character's root pose should constrain global position and heading without imposing a frozen limb posture on the generated motion.

We introduce \textbf{\cwm}, a proxy-conditioned rendering framework comprising a \textbf{bidirectional creation model} and a \textbf{causal four-step streaming model}. Both accept structural proxies, an appearance reference, and language through a shared interface. Training covers different proxy families, such as semantic layouts and shaded geometry, and multiple granularities within a family. Each checkpoint can follow fine constraints when supplied and synthesize plausible local motion when omitted, supporting different input granularities with the same weights.

For video creation, users edit complete proxy sequences to refine camera paths, trajectories, and event timing. The bidirectional model attends across the full clip. During streaming, future states depend on actions yet to occur. The causal model therefore consumes proxy chunks and local text incrementally, generating each chunk in four steps using retained visual history.

We build on MiniMax-H3 \citep{minimaxh3} and adopt Guidance-Free Training (GFT) \citep{chen2025visual} to adapt its guidance-distilled backbone to proxy conditioning. Converting the resulting bidirectional model to a streaming renderer requires both fewer denoising steps and tolerance to errors in generated history. \textbf{Progressive Forcing (ProForcing)} addresses these changes in stages (Figure~\ref{fig:proforcing-pipeline}). Teacher forcing (TF) first trains block-causal prediction on clean video prefixes. Resampling forcing (RF) then constructs imperfect histories by adding noise to observed prefixes and refining them sequentially with the current model.

RF-DMD retains these resampled histories while learning four-step generation through distribution matching distillation (DMD) \citep{yin2024dmd,yin2024dmd2} and an auxiliary GAN loss \citep{jolicoeurmartineau2019relativistic}. The student, trainable score model, and GAN discriminator then continue training on continuous self-generated histories in SF-DMD \citep{huang2025selfforcing,yang2025longlive,ge2026saltpp}. Long-rollout training thus starts with a four-step generator and supervision models already trained alongside it.

We train on an aligned $\sim$200-hour multi-granularity corpus with hierarchical text. Persistent history caching and pipelined proxy encoding and video decoding reduce serving overhead. Across 140 streaming rollouts, \cwm outperforms recent streaming baselines in visual quality, structural adherence, and reference preservation, and maintains stable scene and subject appearance across 30-minute racing rollouts.

Our contributions are:
\begin{itemize}
    \item \textbf{Mixed-granularity proxy conditioning.} A shared conditioning interface lets each creation or streaming checkpoint accept coarse and fine proxies.
    \item \textbf{Progressive Forcing.} A curriculum progresses from TF/RF causal adaptation to four-step DMD with GAN supervision, then continues the trained student, score model, and discriminator under self-generated histories.
    \item \textbf{Long-horizon controllable generation.} Our four-step causal renderer outperforms recent streaming baselines and maintains scene and subject appearance across 30-minute racing rollouts.
    \item \textbf{Multi-granularity corpus and serving stack.} An aligned $\sim$200-hour dataset and a low-latency distributed serving engine support multi-granularity training and continuous interactive streaming.
\end{itemize}

\section{Related Work}
\label{sec:related-work}

\subsection{Interactive Worlds and Generative Rendering}
\label{sec:interactive-worlds}

Action-conditioned video models often internalize world dynamics and visual synthesis in one network, learning latent or discrete action transitions from gameplay and video corpora \citep{bruce2024genie,valevski2024gamengen,decart2024oasis,yu2024gamegenx,wu2024ivideogpt,wu2024pandora,zhang2025matrixgame,mao2025yume}. LingBot-Video studies mixture-of-experts video pretraining for embodied intelligence, combining web videos with robot-oriented footage \citep{ma2026lingbotvideo}. Streaming architectures extend interactive horizons across open-world interactions and explicit context or 3D-aware memory, including the Matrix-Game series, WorldPlay, WorldCrafter, and related streaming world models \citep{he2025matrixgame2,wang2026matrixgame3,qian2026matrixgame35,zhang2026matrixgame4,sun2026worldplay,yu2026worldcrafter,robbyant2026lingbot,gao2026infiniteworlds,yang2026oneira}. Because one visual transition model governs both state evolution and rendering, modifying scene structure or enforcing persistent physical rules remains challenging.

A complementary paradigm maintains explicit state in an executable engine or agentic controller and delegates synthesis to a conditioned video model. Code World Model, Programmable World Model, and AgentGarten compile structured program state into visual signals \citep{chen2026cwm,huang2026pwm,chi2026agentgarten}, Marionette decouples state prediction, geometry rendering, and appearance painting \citep{meng2026marionette}, and Wizard supports real-time world manipulation via agentic control \citep{wu2026wizard}. StreamRender-H3, Streaming CWM, Magpie, and BoxWorld likewise couple game or simulator backends with neural renderers through semantic, white-box, or bounding-box streams \citep{lu2026streamrenderh3,liao2026streamingcwm,zhan2026magpie,wu2026boxworld}. Treating proxies as an explicit state-to-renderer contract, \cwm studies controllable granularity across creation and streaming.

\subsection{Proxy-Conditioned Video Generation}
\label{sec:proxy-conditioned-generation}

Building on spatial control, camera conditioning, and unified video creation frameworks \citep{zhang2023controlnet,he2025cameractrl,jiang2025vace}, proxy-conditioned video models ground synthesis in geometric or semantic structure across several regimes. Paired methods learn directly from dense geometry, lighting, material buffers, or animated meshes, as in DiffusionRenderer, Generative World Renderer, and native 4D mesh rendering \citep{liang2025diffusionrenderer,huang2026generativeworldrenderer,chen2026dar}. Proxy2World relaxes paired supervision via posed RGB-D clips \citep{xu2026proxy2world}, whereas training-free methods such as Code Video Model restrict proxy injection to early denoising steps so appearance follows the reference image \citep{liang2026codevideomodel}.

Recent renderers broaden representation diversity, customization, and structural fidelity. SolarRenderer reuses the backbone's reference pathway across heterogeneous structural videos with a four-step acceleration LoRA \citep{fang2026solarrenderer}, EdenLive and Mundo support real-time reference-driven rendering and customized world synthesis \citep{huang2026edenlive,luo2026mundo}, Fight World Model controls combat interactions via visual proxies \citep{li2026fightworld}, and AnyGameRenderer preserves cross-game entity counts and timing \citep{liu2026anygamerenderer}. \cwm targets an orthogonal axis: structural \emph{granularity} within one representation family, enabling a single model to honor fine constraints when supplied and synthesize natural local motion under lightweight coarse proxies.

\subsection{Streaming Video Diffusion and Few-Step Distillation}
\label{sec:streaming-diffusion}

While training-free queue scheduling such as FIFO-Diffusion enables continuous generation via diagonal denoising \citep{kim2024fifodiffusion}, interactive streaming requires block-causal attention and few-step sampling. Diffusion Forcing trains causal models with independent token noise levels \citep{chen2024diffusionforcing}, CausVid distills a bidirectional teacher into an autoregressive student \citep{yin2025causvid}, and Self Forcing rolls out the student on its own predictions \citep{huang2025selfforcing}. LongLive stabilizes rollouts via streaming tuning and a persistent frame sink \citep{yang2025longlive,xiao2024streamingllm}, SANA-WM and minWM scale minute-horizon and full-stack world modeling \citep{zhu2026sanawm,zhao2026minwm}, and AlayaRenderer-Flash and StreamRender-H3 co-design few-step causal rendering with low-latency serving \citep{lin2026alayaflash,lu2026streamrenderh3}.

To compress flow-matching trajectories \citep{lipman2023flowmatching,liu2023rectifiedflow}, consistency distillation maps states directly to clean targets \citep{song2023consistencymodels}, score-based distribution matching \citep{luo2023diffinstruct,yin2024dmd,zhou2024sid} aligns distributions via auxiliary score differences, and DMD2, ADD, and LADD pair score or diffusion distillation with discriminator supervision \citep{yin2024dmd2,sauer2023add,sauer2024ladd}. Multi-step variants refine supervision across steps, with Trajectory Distribution Matching (TDM) and VA-TDM aligning multi-step trajectories and velocity fields \citep{luo2025tdm,luo2026vatdm} and Endpoint Distribution Matching (eDMD) anchoring the fake score model on final clean outputs \citep{zhao2026edmd}. For causal video, Causal Forcing uses an autoregressive ODE teacher, Causal Forcing++ and Causal-rCM adopt consistency-based causal initialization \citep{zhu2026causalforcing,zhao2026causalforcingpp,zheng2026causalrcm}, and Salt++ aligns streaming generation and scoring contexts \citep{ge2026saltpp}. ProForcing instead distills a four-step student with teacher-feature relativistic GAN supervision \citep{jolicoeurmartineau2019relativistic,sauer2024ladd} on resampled data-anchored histories (RF-DMD) before continuing the coupled student, fake score model, discriminator, and optimizer states on continuous self-generated rollouts (SF-DMD).

\section{Data Curation and Proxy Construction}
\label{sec:data-curation}

Controllable rendering requires aligned training pairs that cleanly separate the geometry prescribed by a proxy from the appearance and local dynamics synthesized by the model. Our corpus comprises approximately 200 hours of footage, combining 150 hours of self-collected data with 50 hours from public datasets, and associates each usable RGB clip with structural conditions, an appearance reference, and hierarchical text annotations. We design the pipeline around three core requirements: spatial-temporal correspondence between proxies and RGB targets, multi-granularity structural representations of the same event, and language descriptions tailored to both whole-window creation and block-causal streaming.

\subsection{Data Sources and Coverage}
\label{sec:data-sources}

\begin{figure}[!t]
  \centering
  \captionsetup{font=small,skip=4pt}
  \includegraphics[width=\linewidth,trim=0 24 0 8,clip]{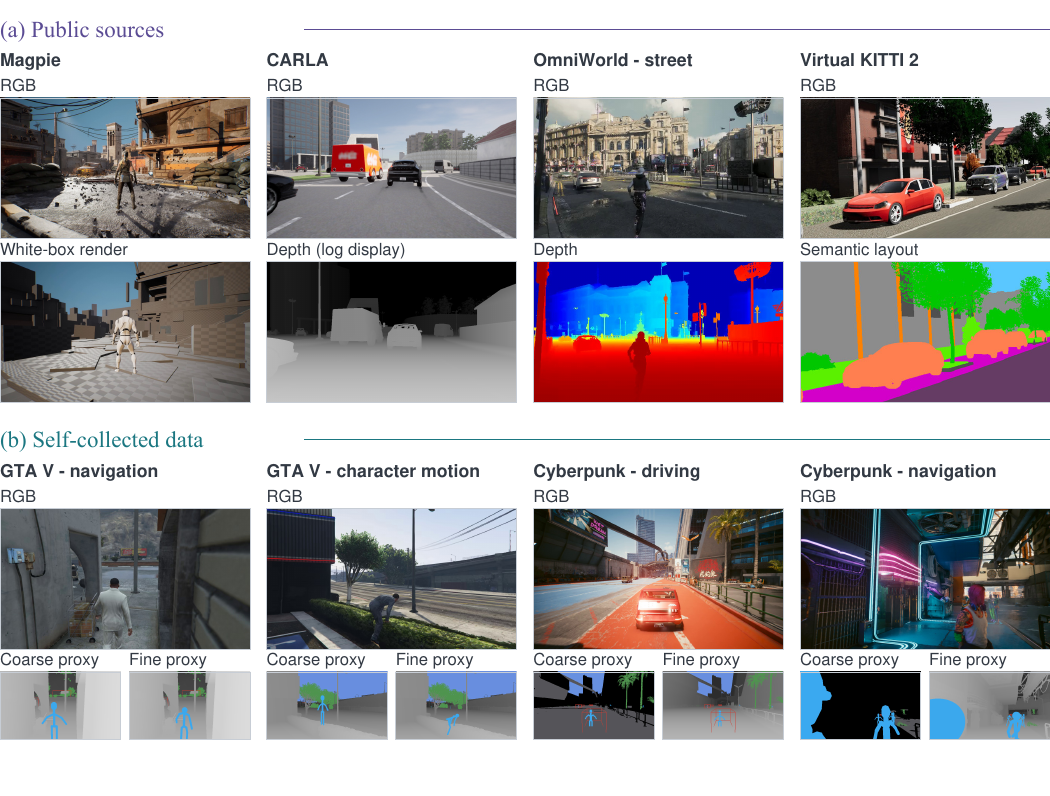}
  \caption{\textbf{Source video frames and aligned structural conditions.} (a) Public datasets pair RGB frames with Magpie white-box geometry, CARLA depth, OmniWorld-Game depth, and Virtual KITTI 2 semantic layouts~\citep{zhan2026magpie,dosovitskiy2017carla,zhou2025omniworld,cabon2020virtualkitti2}. (b) Self-collected GTA V and Cyberpunk 2077 examples pair each RGB source frame with matched coarse and fine proxies of the same event, where coarse proxies place canonical root-driven bodies and fine proxies track articulated limbs.}
  \label{fig:data-sources}
\end{figure}

Table~\ref{tab:data-sources} summarizes the corpus composition. Self-collected recordings allow us to render matched coarse and fine proxies from a shared reconstructed scene state, whereas public datasets broaden environment diversity and contribute native structural modalities. Reported durations count unique source footage prior to extracting overlapping training windows or multiple proxy variants. Figure~\ref{fig:data-sources} illustrates representative source frames alongside their paired structural conditions.

\par\medskip\noindent
\begin{minipage}{\linewidth}
  \centering
  \small
  \captionsetup{font=small,hypcap=false}
  \setlength{\tabcolsep}{5pt}
  \captionof{table}{\textbf{Composition and approximate recording durations of the training corpus.} Public-source hours reflect the subsets used in this work, and all durations count unique source footage before constructing overlapping windows or proxy variants.}
  \label{tab:data-sources}
  \begin{tabular*}{\linewidth}{@{\extracolsep{\fill}}p{0.28\linewidth}rp{0.56\linewidth}@{}}
    \toprule
    Source & Hours & Content and conditioning \\
    \midrule
    Self-collected data & $\sim$150 & Navigation, driving, and character interactions with constructed coarse and fine proxies \\
    OmniWorld~\citep{zhou2025omniworld} & $\sim$15 & Diverse environments and available geometric annotations \\
    Magpie~\citep{zhan2026magpie} & $\sim$15 & Gameplay paired with engine-rendered white-box structure \\
    CARLA~\citep{dosovitskiy2017carla} & $\sim$15 & Driving scenes with semantic layouts and simulator geometry \\
    Other public sources & $\sim$5 & Supplementary video and available structural conditions \\
    \midrule
    Total & $\sim$200 & 150 hours of self-collected data and 50 hours from public sources \\
    \bottomrule
  \end{tabular*}
\end{minipage}\par\medskip

\paragraph{Self-collected data.}
We capture gameplay across diverse 3D environments (including Grand Theft Auto V, Cyberpunk 2077, Elden Ring, and Forza Horizon 5), spanning navigation, urban driving, and character interactions under varied camera trajectories, weather, and illumination. Automated navigation and driving scripts yield sustained, smooth trajectories, while human gameplay contributes unscripted interactions and rapid viewpoint transitions. RGB frames and accessible engine buffers are recorded synchronously, with camera poses and dynamic object states recovered offline where engine exports are incomplete, preserving continuous multi-action trajectories for long-horizon training.

\paragraph{Public sources.}
Public subsets contribute white-box geometry, semantic layouts, and dense depth, and Figure~\ref{fig:data-sources} additionally includes Virtual KITTI 2~\citep{cabon2020virtualkitti2} to illustrate RGB paired with dense semantic segmentation. Source-specific preprocessing aligns native annotations to a unified frame rate and conditioning interface, applying geometric and temporal validity checks before enabling structural supervision. Clips lacking reliable structural alignment are retained solely for reference-and-text appearance supervision. Recording available condition types per sample allows public datasets to contribute their native modalities without requiring synthetic coarse/fine pairs for every source.

\subsection{Aligned Scene States and Multi-Granularity Proxies}
\label{sec:proxy-construction}

\begin{figure}[!t]
  \centering
  \captionsetup{font=small,skip=4pt}
  \includegraphics[width=\linewidth,trim=0 12 0 4,clip]{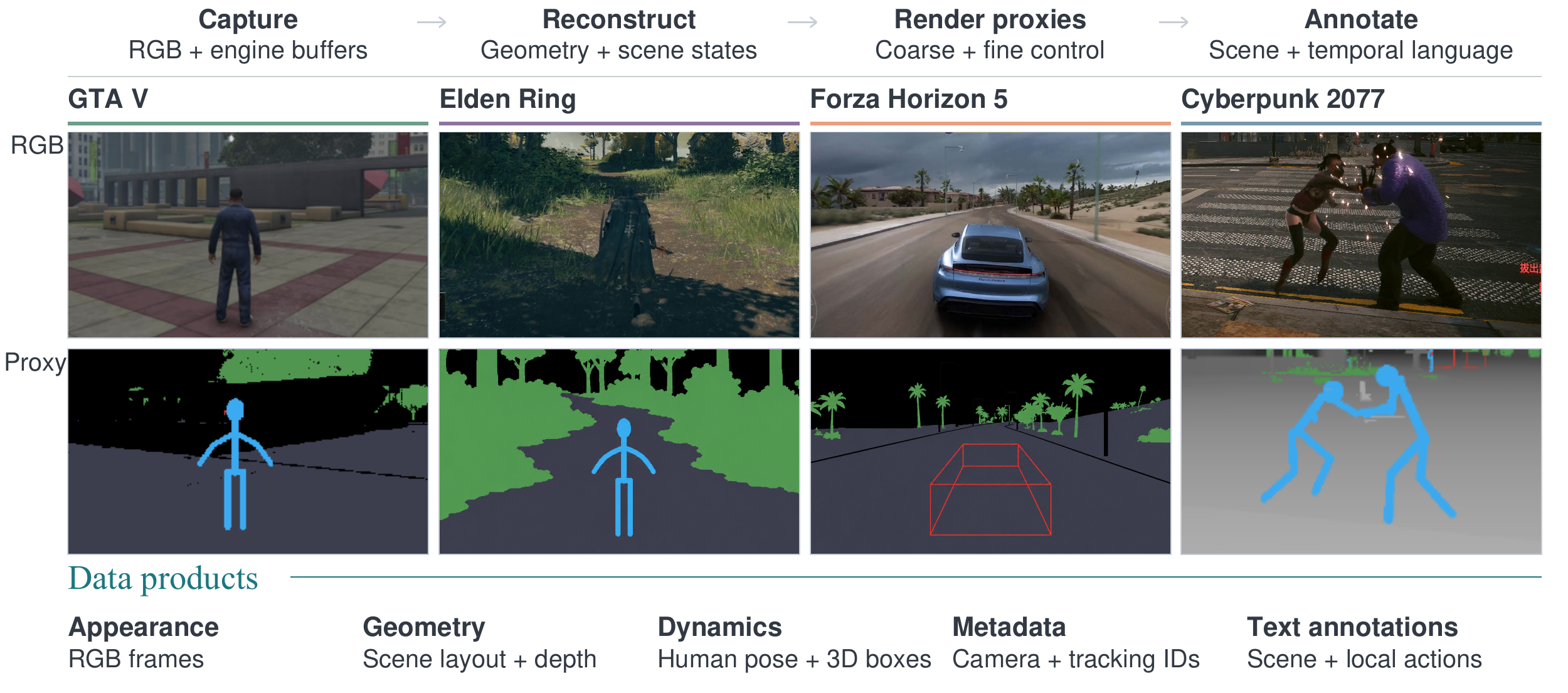}
  \caption{\textbf{Capture and multi-granularity proxy construction pipeline.} Synchronized gameplay capture and 3D scene reconstruction yield aligned appearance targets, geometry, entity dynamics, metadata, and text annotations. From a shared scene state, the pipeline renders matched coarse and fine structural proxies for each event.}
  \label{fig:data-pipeline}
\end{figure}

Figure~\ref{fig:data-pipeline} outlines capture and proxy construction for our self-collected corpus. While recorded RGB serves as the supervision target, synchronized engine buffers and reconstructed 3D scene states (comprising camera intrinsics and extrinsics, static environment geometry, and tracked dynamic entities) allow us to render multiple structural granularities of the exact same visual event.

\paragraph{Capture and reconstruction.}
Starting from synchronized RGB and engine depth capture, our reconstruction pipeline recovers camera calibration, scene semantics, and 3D entity motion. Instance segmentation and multi-object tracking~\citep{ravi2024sam2} isolate moving actors and vehicles from the static scene, which is labeled via semantic segmentation and completed across invalid depth regions~\citep{yang2024depthanythingv2}, while camera pose estimation and background point tracking recover globally consistent trajectories. For dynamic actors, human motion recovery estimates both global root trajectories and articulated body poses, and 3D bounding-box fitting tracks rigid vehicle motion. Retaining per-frame camera parameters, track IDs, and confidence masks alongside the video enables automated reprojection and tracking verification prior to proxy rendering.

\paragraph{Coarse structural control.}
Coarse proxies encode global scene layout, camera motion, and entity placement and heading while omitting fine local geometry. Humans are rendered as canonical rigid mannequins driven solely by root pose, vehicles as oriented 3D boxes, and backgrounds as simplified geometry or semantic regions. Because a canonical mannequin conveys root translation and orientation without limb articulation, the paired RGB target and action text teach the generator to synthesize natural gait and posture consistent with the trajectory.

\paragraph{Fine geometric and motion control.}
Fine proxies augment this representation with articulated human poses (rendered as depth-tested volumetric body parts) and shaded local surface geometry, while preserving rigid vehicle boxes. Wherever both variants are constructed, coarse and fine proxies share identical camera parameters, timestamps, entity identities, and RGB targets, directly supervising the renderer across constraint strengths for the same event. Throughout this work, we distinguish \emph{granularity}, meaning the level of geometric and kinematic detail within a representation, from \emph{representation family}, such as semantic maps, shaded depth, or white-box renders.

\paragraph{Separating structure from appearance.}
To prevent appearance leakage, both granularities strip away surface textures, materials, and scene lighting. Before compositing dynamic entities, foreground masks are excised from background depth and semantic buffers and inpainted from temporally adjacent background views, suppressing silhouette cues from clothing, hair, or accessories. Reprojecting the simplified dynamic primitives onto this cleaned background with depth occlusion ensures that proxies convey pure spatial and motion structure, leaving appearance to the reference frame, language, and learned video prior.

\subsection{Sequence Lengths and Hierarchical Language}
\label{sec:sequence-annotations}

\paragraph{Short windows and continuous sequences.}
Self-collected and public recordings are standardized at 24\,fps into 158-frame windows ($\sim$6.6\,s) and 328-frame continuous sequences ($\sim$13.7\,s). RGB targets use $1344\times768$, while aligned proxies are stored at $672\times384$ and encoded at $672\times384$ or $448\times256$ using nearest-neighbor resizing for semantic maps and anti-aliased filtering for continuous geometry. Short windows support bidirectional GFT, supervised TF/RF, and RF-DMD, whereas 328-frame sequences support variable-length SF-DMD rollouts across uninterrupted trajectories.

\paragraph{Whole-window descriptions.}
Text annotations combine clip-level scene descriptions with timestamped local actions (``Scene + local actions'' in Figure~\ref{fig:data-pipeline}). Each window pairs its opening RGB anchor with a vision-language caption covering environment, subject appearance, lighting, events, and camera motion. The bidirectional creation model and the bidirectional teacher and fake score models each receive independent whole-window captions rather than concatenated local tags.

\paragraph{Streaming action annotations.}
Timestamped local annotations track action progression, motion directions, interactions, and camera shifts across block boundaries so that TF/RF and the distilled student observe text strictly up to the current block, without future labels or clip summaries. For coarse proxies, local text disambiguates behaviors whose articulation is omitted from rigid root geometry.

\paragraph{Condition organization.}
A system instruction defines the roles of structural proxies and the appearance reference alongside sample captions, while automated validation checks reference tags, temporal order, and entity-action agreement.

\subsection{Quality Control and Evaluation Separation}
\label{sec:data-quality}

Quality control screens both video and proxies for corrupted or frozen frames, exposure failures, shot cuts, reprojection outliers, tracking switches, and timing drift. Automated geometric checks and manual review evaluate each proxy against its intended constraints, distinguishing omitted coarse articulation from root-trajectory or camera errors.

Records track source identifiers, frame ranges, proxy variants, captions, and encoding provenance, with cross-source deduplication across shared game worlds. Training and held-out splits are partitioned by recording (and by scene or map where identifiable), keeping overlapping windows, coarse/fine pairs, and alternative views of one recording in the same split to prevent leakage.

\Needspace{8\baselineskip}
\section{Invoker-Render}
\label{sec:method}

\cwm adapts MiniMax-H3~\citep{minimaxh3} into both a bidirectional video creation model and a causal four-step streaming renderer. We first formalize the shared conditioning interface and linear flow parameterization, then present bidirectional proxy adaptation via Guidance-Free Training, and finally detail Progressive Forcing (ProForcing), which distills few-step causal rendering on data-anchored histories before transitioning to continuous self-generated rollouts.

\subsection{Rendering Formulation and Proxy Conditioning}
\label{sec:rendering}

Let $\bm{x}_0=(\bm{x}_0^1,\ldots,\bm{x}_0^B)$ denote clean video latents partitioned into $B$ contiguous temporal blocks (corresponding to streaming chunks at inference time). Conditioning inputs consist of an independently encoded appearance anchor $a$, structural proxies $p=p_{1:B}$, a system instruction $s$, and text annotations $\ell=\ell_{1:B}$. Whereas the bidirectional creation model attends over the full proxy sequence $p_{1:B}$ and a clip-level caption, the causal streaming model observes $c_b=(a,s,p_{\leq b},\ell_{\leq b})$ at block $b$ and factorizes the joint distribution autoregressively across blocks:
\begin{equation}
\label{eq:causal-factorization}
p_\theta(\bm{x}_0\mid a,p,s,\ell)
=\prod_{b=1}^{B}p_\theta(\bm{x}_0^b\mid\bm{h}^b,c_b),
\end{equation}
where $\bm{h}^b$ denotes the retained visual history prior to block $b$ and $\ell_{\leq b}$ collects text annotations revealed through block $b$. Target attention in the streaming model is bidirectional within each block and strictly causal across blocks. Both models preserve the Diffusion Transformer (DiT)~\citep{peebles2023dit} backbone's native reference pathway and rotary positional embedding (RoPE)~\citep{su2024roformer} indexing, keeping the appearance anchor $a$ fixed outside the training loss.

\paragraph{Control across granularities.}
Coarse and fine proxies enter through an identical conditioning pathway, trained jointly on mixed-granularity examples. Coarse inputs prescribe scene layout, camera trajectory, and entity root motion, while fine inputs further constrain limb articulation and local surface geometry. At inference time, downstream applications can select the structural detail suited to each interaction while relying on the appearance reference, text, and video prior to supply visual texture and unconstrained local motion.

\paragraph{Flow parameterization.}
Given clean latents $\bm{x}_0$, Gaussian noise $\bm{\epsilon}\sim\mathcal{N}(\bm{0},\bm{I})$, and noise level $t\in(0,1]$, we adopt the linear flow-matching trajectory and velocity target~\citep{lipman2023flowmatching,liu2023rectifiedflow}
\begin{equation}
\label{eq:flow}
\bm{x}_t=(1-t)\bm{x}_0+t\bm{\epsilon},
\qquad \bm{v}=\bm{\epsilon}-\bm{x}_0.
\end{equation}
We write $\bm{v}_\theta$ for the velocity network and $\bm{f}_\theta=\bm{x}_t-t\bm{v}_\theta$ for the induced clean-latent denoiser, omitting shared conditioning arguments where unambiguous. Here $\theta^-$ indicates evaluation at current weights with gradients detached, and all squared-error objectives below are implicitly averaged over valid non-anchor latent elements.

\subsection{Bidirectional Creation Model}
\label{sec:creation}

Because MiniMax-H3 already incorporates distilled classifier-free guidance~\citep{ho2022cfg}, we adapt it to structural proxy control using Guidance-Free Training (GFT)~\citep{chen2025visual}. Pairing the full conditioning tuple $c^+$ with a weakened counterpart $c^-$ that drops the structural proxy while retaining the appearance anchor, we blend their velocity predictions at guidance scale $\beta=2$ with 10\% condition dropout:
\begin{equation}
\label{eq:gft}
\begin{gathered}
\widetilde{\bm{v}}_\theta
=\frac{1}{\beta}\bm{v}_\theta(\bm{x}_t,t\mid c^+)
+\left(1-\frac{1}{\beta}\right)\bm{v}_{\theta^-}(\bm{x}_t,t\mid c^-),\\[3pt]
\mathcal{L}_{\mathrm{GFT}}(\theta)
=\mathbb{E}_{\bm{x}_0,\bm{\epsilon},t}\!\left[
w(t)\left\lVert\widetilde{\bm{v}}_\theta-\bm{v}\right\rVert_2^2
\right].
\end{gathered}
\end{equation}
Here $w(t)$ denotes the backbone's time-dependent loss weighting. Detaching the weak-condition branch $\bm{v}_{\theta^-}(\bm{x}_t,t\mid c^-)$ routes gradients exclusively through the proxy-conditioned pathway, sharpening structural adherence without inference-time CFG overhead. Trained under full bidirectional attention with complete proxy sequences and whole-window captions, the resulting checkpoint serves directly as our offline creation renderer and provides the initialization for causal adaptation and distillation critics.

\subsection{Progressive Forcing for Few-Step Rendering}
\label{sec:proforcing}

\begin{figure}[!t]
  \centering
  \captionsetup{font=small,skip=4pt}
  \includegraphics[width=\linewidth]{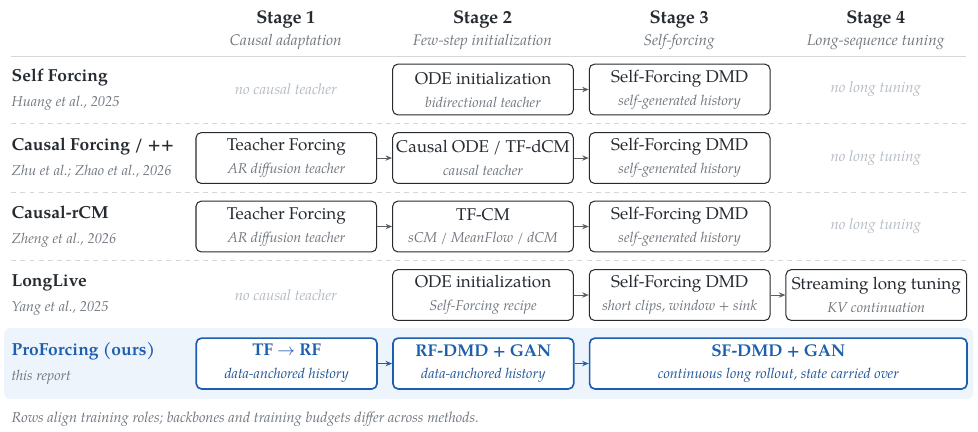}
  \caption{\textbf{Comparison of ProForcing with representative causal video training recipes.}
  Rows align five training pipelines across four functional stages from causal adaptation (Stage~1) to long-sequence tuning (Stage~4).
  Self Forcing initializes from a bidirectional ODE teacher~\citep{huang2025selfforcing}, Causal Forcing/++ and Causal-rCM initialize from causal teachers~\citep{zhu2026causalforcing,zhao2026causalforcingpp,zheng2026causalrcm}, and LongLive appends streaming long tuning~\citep{yang2025longlive}.
  ProForcing adapts block-causal attention via TF and RF (Stage~1), distills four-step RF-DMD with GAN supervision under data-anchored histories (Stage~2), and transfers the student, fake score model, discriminator, optimizer states, and student exponential moving average (EMA) directly into continuous long-rollout SF-DMD (Stages~3--4).}
  \label{fig:proforcing-pipeline}
\end{figure}

Directly distilling a bidirectional model into a few-step autoregressive renderer conflates trajectory compression with exposure to self-generated rollout errors. ProForcing disentangles these challenges across progressive stages (Figure~\ref{fig:proforcing-pipeline}). Supervised TF and RF first adapt attention to block-causal histories anchored to ground-truth video, RF-DMD then distills a four-step generator alongside learned score and adversarial critics under those data-anchored histories, and SF-DMD (Section~\ref{sec:self-forcing}) finally transitions the jointly trained system to continuous self-generated rollouts.

\Needspace{8\baselineskip}
\paragraph{Causal adaptation with TF and RF.}
Replacing bidirectional target attention with block-causal attention, both Teacher Forcing (TF) and Resampling Forcing (RF) optimize the GFT objective of Eq.~\eqref{eq:gft} under $\beta=2$ and 10\% condition dropout. TF first conditions each block on clean ground-truth history $\bm{h}_{\mathrm{TF}}^b=\bm{x}_0^{<b}$. RF then bridges clean prefixes and imperfect inference context by corrupting each observed history block to noise level $\rho$ and refining it sequentially with the current detached model:
\begin{equation}
\label{eq:rf-history}
\begin{aligned}
\widetilde{\bm{x}}_\rho^b
&=(1-\rho)\bm{x}_0^b+\rho\bm{\epsilon}^b,\\[3pt]
\widetilde{\bm{x}}_0^b
&=\bm{f}_{\theta^-}\!\left(
\widetilde{\bm{x}}_\rho^b,\rho\mid\widetilde{\bm{x}}_0^{<b},c_b
\right).
\end{aligned}
\end{equation}
Conditioning each one-step refinement on previously refined blocks $\widetilde{\bm{x}}_0^{<b}$ yields $\bm{h}_{\mathrm{RF}}^b=\widetilde{\bm{x}}_0^{<b}$, exposing the model to realistic reconstruction errors while remaining anchored to observed trajectories. Throughout both TF and RF, history construction is detached, proxies and streaming text arrive block by block under the same $\beta=2$ GFT supervision and 10\% dropout, and loss is computed across all target blocks.

\paragraph{RF-conditioned distribution matching.}
Initialized from the supervised RF checkpoint, RF-DMD compresses sampling to four steps via distribution matching distillation~\citep{yin2024dmd,yin2024dmd2,luo2025tdm,zhao2026edmd} while retaining detached RF history construction ($\bm{h}_{\mathrm{RF}}^b$). Keeping histories anchored to observed video during distillation isolates four-step trajectory learning from open-loop autoregressive drift. To score student outputs, a frozen bidirectional teacher $\bm{f}_{\mathrm{teacher}}$ and a trainable bidirectional fake score model $\bm{f}_{\mathrm{fake}}$ (both initialized from the bidirectional GFT checkpoint) evaluate re-noised student predictions under complete scoring-window conditions, while the student itself remains strictly block-causal. Meanwhile, $\bm{f}_{\mathrm{fake}}$ is updated online via flow matching on detached, re-noised student samples.

Specifically, let $\bm{x}_0^\theta$ denote the clean prediction tapped from a selected step of the student's four-step denoising rollout, and let $\bm{x}_t=(1-t)\bm{x}_0^{\theta^-}+t\bm{\epsilon}$ be its detached re-noised input shared by both score models. We apply the DMD gradient through the surrogate objective
\begin{equation}
\label{eq:dmd}
\mathcal{L}_{\mathrm{DMD}}(\theta)
=\frac12\,\mathbb{E}\!\left[
\left\lVert\bm{x}_0^\theta-\bm{x}_0^{\theta^-}+\bm{g}\right\rVert_2^2
\right].
\end{equation}
Writing $\bm{g}=(\bm{g}^1,\ldots,\bm{g}^B)$, the target-seeking displacement on block $b$ is $\bm{g}^b=(\bm{f}_{\mathrm{fake}}^b-\bm{f}_{\mathrm{teacher}}^b)/Z_b$, where $Z_b$ normalizes by the mean absolute discrepancy between detached student and teacher predictions (clamped to a positive floor). Treating $\bm{g}$ as a stop-gradient target ensures that differentiation of Eq.~\eqref{eq:dmd} propagates the exact normalized score difference into $\bm{x}_0^\theta$ across all valid blocks.

\paragraph{Teacher-feature relativistic GAN.}
To sharpen perceptual detail and suppress few-step over-smoothing, we pair DMD with a discriminator $D_\phi$ comprising frozen teacher backbone features~\citep{sauer2024ladd,yin2024dmd2} followed by lightweight trainable projections and a noise-conditioned scalar head. Ground-truth windows $\bm{x}_0$ and generated windows $\bm{x}_0^\theta$ are perturbed with identical $(t,\bm{\epsilon})$ via Eq.~\eqref{eq:flow} under matched conditions and the shared appearance anchor, producing scalar logits $r_\phi$ and $q_\phi$. We optimize paired relativistic logistic losses~\citep{jolicoeurmartineau2019relativistic,yin2024dmd2} alongside the combined student objective:
\begin{equation}
\label{eq:student}
\begin{aligned}
\mathcal{L}_{D}(\phi)
&=\lambda_D\,\mathbb{E}\!\left[\log\!\left(1+e^{q_\phi-r_\phi}\right)\right],\\[3pt]
\mathcal{L}_{\mathrm{GAN}}(\theta)
&=\mathbb{E}\!\left[\log\!\left(1+e^{\operatorname{sg}(r_\phi)-q_\phi}\right)\right],\\[3pt]
\mathcal{L}_{\mathrm{student}}(\theta)
&=\mathcal{L}_{\mathrm{DMD}}(\theta)
+\lambda_{\mathrm{adv}}\mathcal{L}_{\mathrm{GAN}}(\theta).
\end{aligned}
\end{equation}
Here $\lambda_D$ and $\lambda_{\mathrm{adv}}$ weight the discriminator and adversarial terms, and $\operatorname{sg}(\cdot)$ denotes the stop-gradient operator on scalar logits. Discriminator updates detach generated latents $\bm{x}_0^\theta$, whereas student updates detach the real-branch logit $r_\phi$ and freeze both teacher and discriminator parameters while backpropagating through the generated-branch input $q_\phi$. Note that $D_\phi$ and $\bm{f}_{\mathrm{fake}}$ remain separate networks trained under distinct objectives.

\subsection{Long-Sequence Self-Forcing}
\label{sec:self-forcing}

Building on self-forcing and streaming long tuning~\citep{huang2025selfforcing,yang2025longlive}, SF-DMD replaces data-anchored RF prefixes with continuous student-generated histories. Starting from the appearance anchor $a$, the student sequentially denoises each block in four steps and commits its detached terminal prediction $\widehat{\bm{x}}_0^b$ to the causal history cache ($\bm{h}_{\mathrm{SF}}^b=\widehat{\bm{x}}_0^{<b}$). While detached terminal outputs advance the rollout state, selected step predictions $\bm{x}_0^\theta$ retain local computation graphs for DMD and GAN supervision. Student weights remain fixed throughout the multi-block rollout, accumulate gradients across scored intervals, and take a single consolidated update before the next rollout refreshes its history from scratch.

\paragraph{Local scoring of a continuous stream.}
To supervise rollouts longer than the bidirectional teacher's native window, we tile the continuous student stream with overlapping local scoring windows. Each window pairs preceding generated blocks (serving purely as detached temporal context) with a disjoint interval of newly generated target blocks, always retaining the original appearance anchor $a$. The teacher, fake score model, and discriminator observe complete conditions within each local window while the student remains causal, and DMD and GAN input gradients are masked to apply strictly once per non-anchor target block, weighted by target coverage.

\paragraph{Coordinated continuation from RF-DMD.}
Crucially, transitioning from RF-DMD to SF-DMD carries forward the entire coupled training state (student weights, fake score model $\bm{f}_{\mathrm{fake}}$, discriminator $D_\phi$, their respective AdamW moments~\citep{loshchilov2019adamw}, and the student EMA) against the same frozen teacher. Because RF-DMD already converges a capable four-step causal generator together with calibrated supervision models tailored to its output distribution, SF-DMD avoids cold-start critic instability and focuses optimization on mitigating long-horizon error accumulation under self-generated histories.

\section{Streaming Serving Infrastructure}
\label{sec:infrastructure}

\begin{figure}[!t]
  \centering
  \captionsetup{font=small,skip=4pt}
  \includegraphics[width=\linewidth]{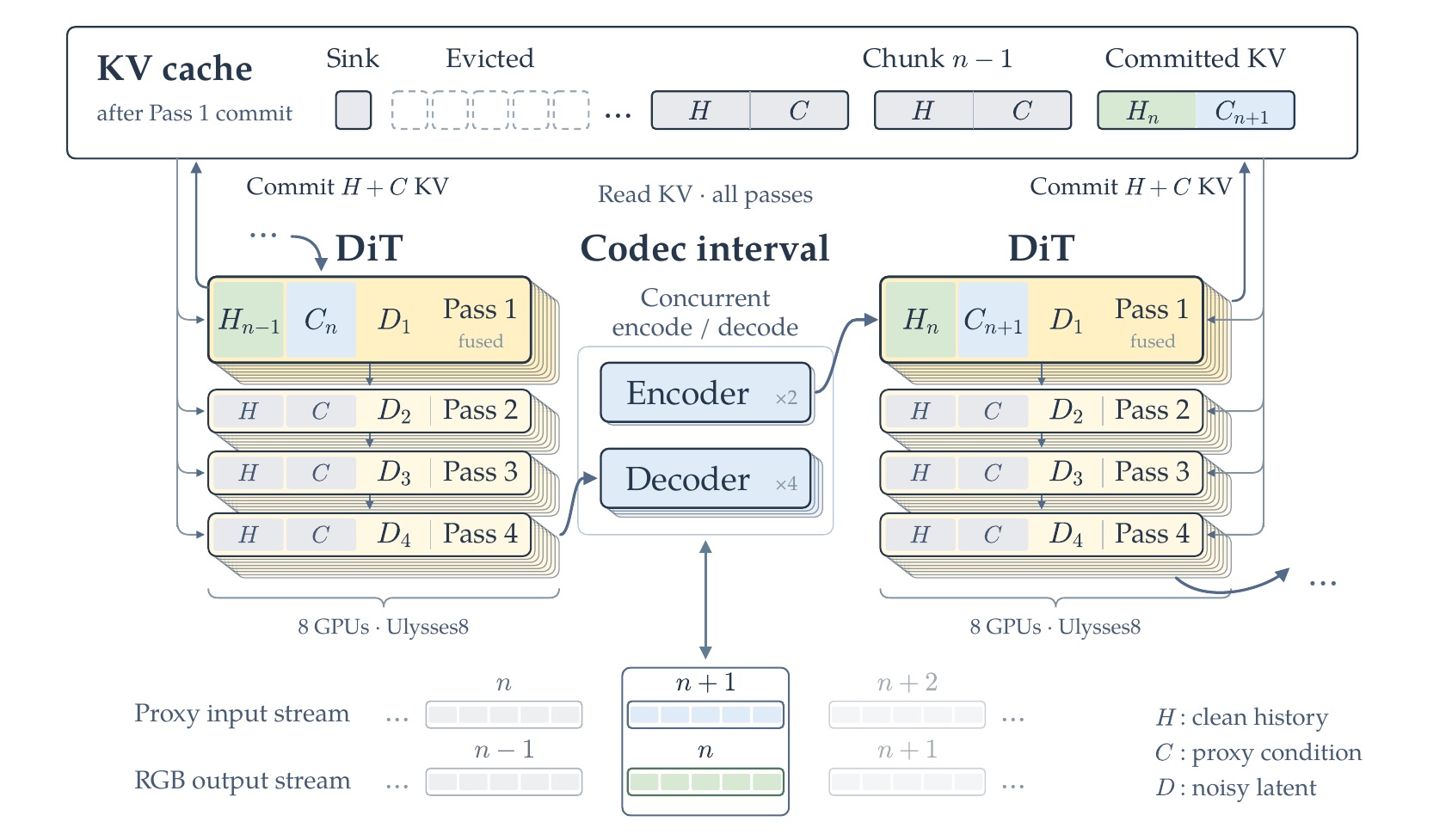}
  \caption{\textbf{Streaming inference schedule on an eight-GPU colocated pool.}
  Each ordinary continuation chunk runs four DiT passes under Ulysses-8 context parallelism.
  Pass~1 fuses deferred clean history $H_{n-1}$, the current proxy condition $C_n$, and the first denoising step $D_1$; Passes~2--4 finish denoising, and only clean $H{+}C$ key/value states are committed.
  During the codec interval between DiT windows, proxy encoding for chunk $n{+}1$ overlaps with causal video decoding ($\times 4$) for chunk $n$.}
  \label{fig:streaming-schedule}
\end{figure}

To deploy the causal world model as an online interactive rendering service, each streaming step must encode incoming visual proxies, update retained clean history $H$ and current proxy conditioning $C$ within a bounded KV window, run four-step causal denoising, and return decoded RGB frames within a strict latency budget.
Starting from a standard streaming baseline, we streamline the inference stack across three levels:
(i)~an \textbf{asymmetric streaming codec} pairing a lightweight Tiny AutoEncoder (TAE) for online proxy encoding with a causal video decoder for RGB reconstruction;
(ii)~\textbf{intra-window invariant reuse and fused execution} across DiT layers, KV workspaces, and prefill/denoising passes; and
(iii)~\textbf{cross-packet codec overlap and asynchronous CPU delivery} (Figure~\ref{fig:streaming-schedule}).
On an eight-GPU node at $1344\times 768$ resolution, these optimizations reduce the steady-state per-chunk period from \textbf{1770~ms to 930~ms} and increase throughput from \textbf{9.9~FPS to 18.0~FPS} (\textbf{1.82$\times$} speedup; Figure~\ref{fig:streaming-performance}).

\subsection{Efficient Asymmetric Streaming Codec}
\label{sec:streaming-codec}

Interactive streaming places asymmetric demands on the visual codec: incoming proxy frames must be encoded at every chunk with minimal latency, whereas generated latents must be decoded into temporally continuous RGB video.
Encoding online proxies with the full video encoder is slower than output decoding and dominates the inter-chunk codec interval.

We therefore adopt an asymmetric codec design.
The static reference frame is encoded once at session start with the full encoder to preserve anchor detail, while online proxy packets are encoded independently by a lightweight \textbf{TAE proxy encoder} directly into normalized $16\times$ spatially compressed, 24-channel conditioning latents.
On the output side, we retain the tensor-parallel \textbf{Causal Video Decoder} to preserve temporal continuity across chunks.
Replacing the online proxy encoder with TAE makes input encoding an order of magnitude faster than causal decoding ($T_{\mathrm{enc}} \ll T_{\mathrm{dec}}$), reducing the steady-state chunk period from \textbf{1770~ms to 1410~ms} and raising throughput from \textbf{9.9~FPS to 12.4~FPS}.

\subsection{Intra-Window Reuse and Fused Execution}
\label{sec:inference-reuse}

Within each DiT window, we eliminate redundant per-layer preparation, projection compute, memory allocation, and separate prefill launches:

\begin{itemize}
    \item \textbf{Cross-layer structural sharing.}
    Within a single forward pass, all DiT layers share identical token layout plans, sliding-window crop indices, and frame-aligned RoPE position tensors.
    We construct these plans and index tensors once at backbone entry and reuse them across all layers.

    \item \textbf{Per-layer AdaLN projection caching.}
    Because session conditioning is stationary and the four-step Euler timestep schedule repeats across chunks, we cache each DiT block's projected AdaLN modulation parameters by timestep and reuse them across subsequent chunks.

    \item \textbf{Transactional KV workspace reuse.}
    Instead of allocating fresh output and concatenation buffers at every layer, we reuse pre-allocated KV result backplanes and temporary forward workspaces across the transaction lifecycle (Figure~\ref{fig:streaming-schedule}, top).
    Combining structural invariant and workspace reuse reduces the steady-state chunk period from \textbf{1410~ms to 1180~ms} and lifts throughput from \textbf{12.4~FPS to 14.4~FPS}.

    \item \textbf{Fused prefill-and-denoise pass ($\text{H/C/D}_1$).}
    In ordinary continuation windows, we defer committing prior clean history $H_{n-1}$ and pack it with incoming proxy tokens $C_n$ and the first denoising state $D_1$ into a single block-causal composite forward (Pass~1 in Figure~\ref{fig:streaming-schedule}), followed by three denoising-only passes ($D_2$--$D_4$).
    Reducing backbone launches from six to four trims the chunk period from \textbf{1180~ms to 1165~ms} and raises throughput to \textbf{14.8~FPS}.
\end{itemize}

\subsection{Cross-Packet Pipelining and RGB Delivery}
\label{sec:codec-pipelining}

With intra-window DiT execution compressed, we pipeline codec execution and host frame delivery across packet boundaries to remove inter-chunk idle gaps:

\begin{itemize}
    \item \textbf{Cross-packet codec overlap.}
    Between consecutive DiT windows, causal video decoding for chunk $n$ and TAE proxy encoding for chunk $n{+}1$ run concurrently on disjoint GPU subsets (Figure~\ref{fig:streaming-schedule}, center and bottom).
    Because TAE encoding finishes well within the causal decoding window, next-packet proxy encoding is hidden behind current-packet decoding.

    \item \textbf{Pipelined CPU delivery.}
    Once causal decoding and the device-to-host copy into shared RGB memory complete, frame packing and socket emission are offloaded to an asynchronous CPU worker.
    The GPU pool immediately starts chunk $n{+}1$'s DiT passes while host delivery proceeds in parallel, cutting the steady-state chunk period from \textbf{1165~ms to 930~ms} and reaching \textbf{18.0~FPS}.
\end{itemize}

\par\medskip\noindent
\begin{minipage}{\linewidth}
  \centering
  \captionsetup{font=small,skip=4pt,hypcap=false}
  \begin{minipage}{0.54\linewidth}
    \centering
    \includegraphics[width=\linewidth]{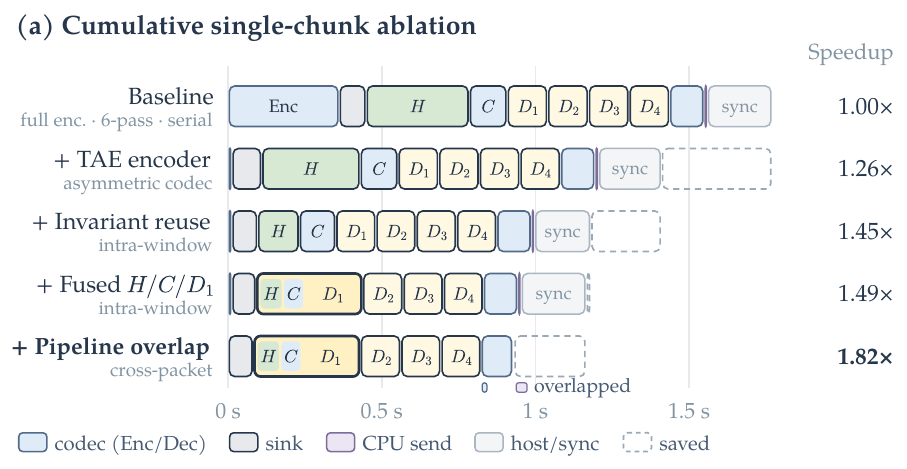}
  \end{minipage}%
  \begin{minipage}{0.46\linewidth}
    \centering
    \includegraphics[width=\linewidth]{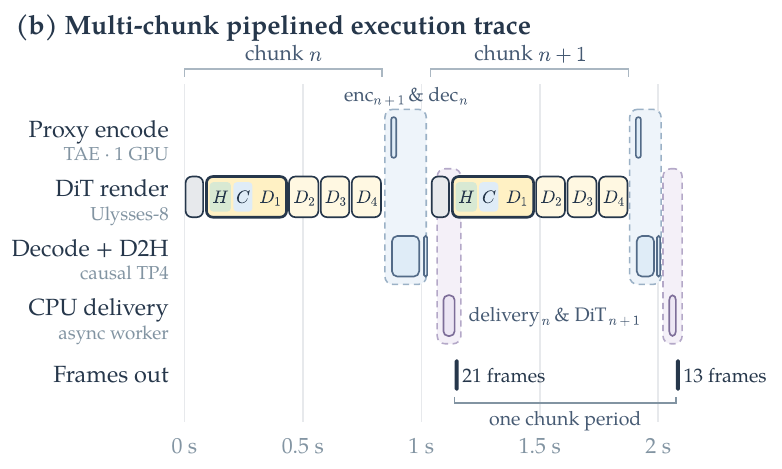}
  \end{minipage}
  \captionof{figure}{\textbf{Streaming inference performance.} (a) Cumulative per-chunk latency ablation showing the reduction in execution path and codec overhead. (b) Multi-chunk pipeline trace demonstrating concurrent codec and asynchronous delivery execution.}
  \label{fig:streaming-performance}
\end{minipage}\par\medskip

\section{Experiments}
\label{sec:experiments}

We evaluate \cwm through systematic pairwise preference comparisons on causal streaming rollouts alongside qualitative studies of multi-granularity creation, camera control, and long-horizon streaming. Section~\ref{sec:experimental-setup} details the held-out benchmark and evaluation protocol, Section~\ref{sec:quantitative-results} compares our four-step causal renderer against representative streaming baselines, and Section~\ref{sec:qualitative-results} illustrates control across proxy families, structural granularities, camera paths, and extended rollouts.

\subsection{Experimental Setup}
\label{sec:experimental-setup}

\paragraph{Benchmark and baselines.}
We evaluate causal streaming rendering on a held-out benchmark of 140 test sequences spanning 6.58\,s to 20\,s across diverse 3D environments, multi-entity interactions, and proxy representation families, covering both single-window clips and multi-window streaming rollouts. We compare our ProForcing-trained \textbf{\cwm (Causal, 4-step)} checkpoint against three representative few-step causal and streaming proxy-conditioned renderers: \textbf{SolarRenderer (Causal, 4-step)}~\citep{fang2026solarrenderer}, \textbf{StreamRender-H3 (2-step)}~\citep{lu2026streamrenderh3}, and \textbf{Streaming CWM (4-step)}~\citep{liao2026streamingcwm}. Every method receives the identical structural proxy video and appearance reference image on each test sequence.

\paragraph{Pairwise preference protocol and Elo metrics.}
Effective streaming proxy rendering demands sharp per-frame synthesis, faithful spatial-temporal alignment to the driving proxy, and sustained reference fidelity across sliding windows. Following SolarRenderer~\citep{fang2026solarrenderer}, we use \textbf{Gemini 3.8 Flash} to conduct blinded pairwise evaluations along four criteria. Each query presents the ground-truth structural proxy sequence, the appearance reference image, and two anonymized candidate renderings $A$ and $B$ (uniformly sampled as timestamped frame sequences at $672\times384$ and 12\,fps with randomized order balancing), eliciting an independent win, tie, or loss judgment along four axes:
\begin{itemize}
    \item \textbf{Visual Quality:} per-frame detail sharpness, suppression of synthesis artifacts, and temporal smoothness.
    \item \textbf{Structure \& Motion:} adherence of scene geometry, subject trajectories, action timing, and camera motion to the input proxy.
    \item \textbf{Reference Preservation:} stability of palette, illumination, material appearance, and subject identity relative to the reference frame.
    \item \textbf{Overall Preference:} holistic quality as a coherent, deliverable rendered sequence.
\end{itemize}
Pairwise preferences are summarized as \textbf{Elo ratings} using the Bradley--Terry model~\citep{bradley1952rank}, with ties counted as half a win and the mean rating across methods set to 1000. Overall Preference is evaluated independently of the three component criteria.

\subsection{Quantitative Streaming Comparison}
\label{sec:quantitative-results}

\begin{figure}[!htb]
  \centering
  \captionsetup{font=small,skip=4pt}
  \includegraphics[width=\linewidth]{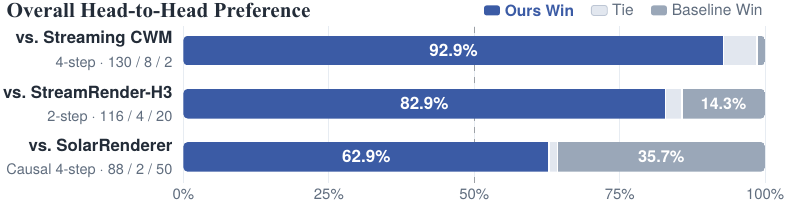}
  \caption{\textbf{Head-to-head overall preference against streaming baselines.} Bars show pairwise win, tie, and loss proportions of \cwm (Causal, 4-step) across 140 blinded test comparisons per baseline, with the dashed vertical line serving as a 50\% reference line.}
  \label{fig:streaming-winrate}
\end{figure}

Table~\ref{tab:streaming-elo} reports Elo ratings across four criteria, and Figure~\ref{fig:streaming-winrate} shows Overall outcomes on 140 sequences per baseline. \cwm ranks first on all four criteria, with an Overall rating of \textbf{1232.0}.

\par\medskip\noindent
\begin{minipage}{\linewidth}
  \centering
  \small
  \captionsetup{font=small,hypcap=false}
  \setlength{\tabcolsep}{5.5pt}
  \captionof{table}{\textbf{Four-dimensional Elo ratings for causal and streaming proxy-conditioned rendering.} Results are evaluated on 140 held-out test sequences, with the mean Elo rating normalized to 1000 ($\uparrow$ higher is better). Best results are in \textbf{bold}.}
  \label{tab:streaming-elo}
  \begin{tabular}{@{}lccccc@{}}
    \toprule
    Method & Steps & Visual Quality $\uparrow$ & Structure \& Motion $\uparrow$ & Ref.\ Preservation $\uparrow$ & Overall $\uparrow$ \\
    \midrule
    Streaming CWM~\citep{liao2026streamingcwm} & 4-step & 653.0 & 626.7 & 835.3 & 692.4 \\
    StreamRender-H3~\citep{lu2026streamrenderh3} & 2-step & 935.7 & 1011.4 & 882.5 & 940.2 \\
    SolarRenderer (Causal)~\citep{fang2026solarrenderer} & 4-step & 1174.9 & 1152.9 & 1118.1 & 1135.3 \\
    \midrule
    \textbf{\cwm (Causal, Ours)} & \textbf{4-step} & \textbf{1236.4} & \textbf{1209.0} & \textbf{1164.1} & \textbf{1232.0} \\
    \bottomrule
  \end{tabular}
\end{minipage}\par\medskip

Against Streaming CWM, \cwm records 130 wins, 8 ties, and 2 losses, giving a 92.9\% win rate and an Overall margin of $+539.6$ Elo. Against StreamRender-H3, it records 116 wins, 4 ties, and 20 losses (82.9\% wins and $+291.8$ Overall Elo). Its margins over StreamRender-H3 are $+300.7$ in Visual Quality, $+197.6$ in Structure \& Motion, and $+281.6$ in Reference Preservation.

SolarRenderer (Causal) is the strongest baseline across all four criteria. \cwm leads by $+61.5$ in Visual Quality, $+56.1$ in Structure \& Motion, and $+46.0$ in Reference Preservation. Its 88 wins, 2 ties, and 50 losses against SolarRenderer (62.9\% wins vs.\ 35.7\% losses) correspond to an Overall advantage of \textbf{$+96.7$ Elo}.

\begin{figure}[!htb]
  \centering
  \captionsetup{font=small,skip=4pt}
  \includegraphics[width=\linewidth]{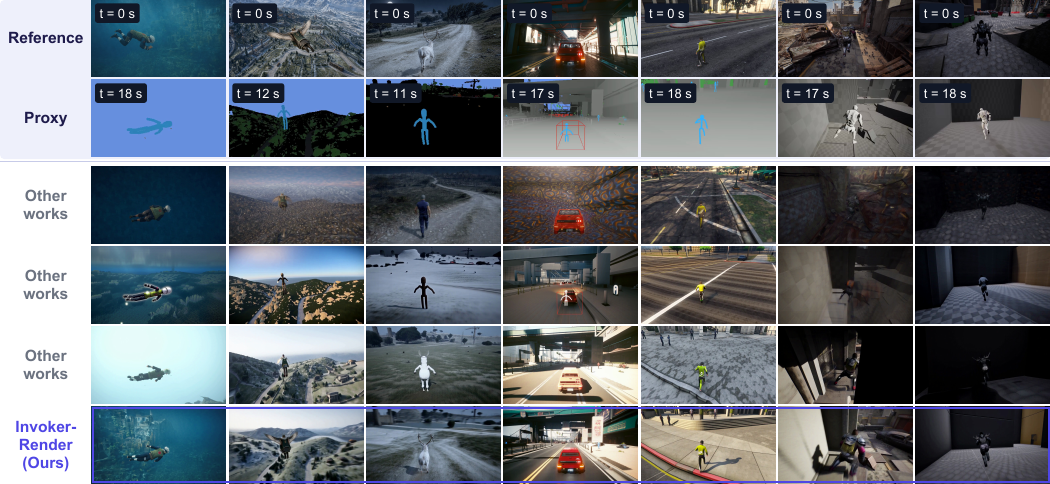}
  \caption{\textbf{Qualitative comparison of causal streaming renderers on held-out rollouts.} Each column shows one test sequence at a matched late-rollout timestamp (11\,s to 18\,s). The top shaded rows display the 0\,s appearance reference and the synchronized input proxy, followed by frames from other works and \cwm (outlined at bottom), which maintains both structural fidelity and reference appearance at the shown late-rollout timestamps.}
  \label{fig:causal-comparison}
\end{figure}

Figure~\ref{fig:causal-comparison} illustrates seven held-out sequences at matched late-rollout timestamps. Other works exhibit scene-geometry loss, severe texture artifacts, visible proxy geometry and flat proxy colors, or drift from the appearance reference at longer horizons: the underwater scene shifts into a bright sky, the winged character loses its wings, the deer loses its antlers, the red car turns beige, the road surface becomes blotchy, half of the ruined street turns black, and a narrow alley fades almost completely to black. Across all seven sequences, \cwm preserves both structural alignment and reference identity.

\subsection{Qualitative Results}
\label{sec:qualitative-results}

\begin{figure}[!htb]
  \centering
  \captionsetup{font=small,skip=4pt}
  \includegraphics[width=\linewidth]{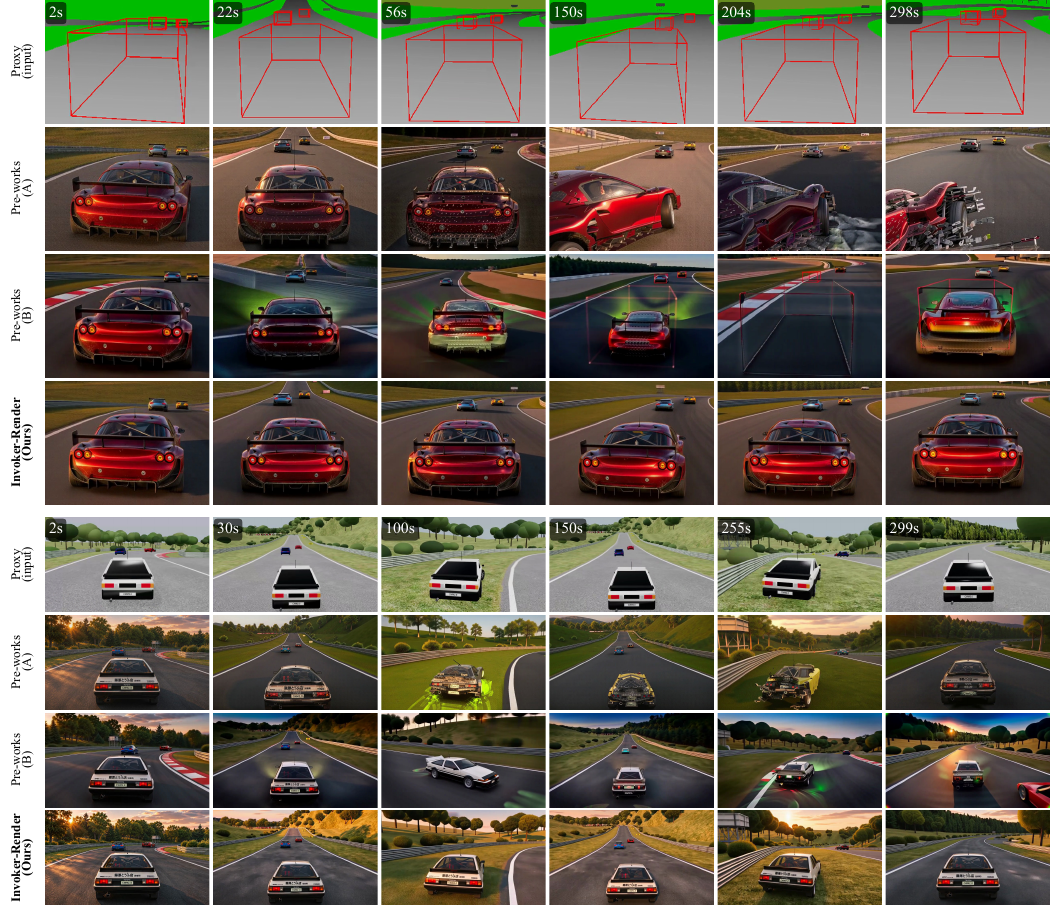}
  \caption{\textbf{Long-horizon causal rendering on two racing sequences.}
  Each case shows six matched timestamps, read from left to right. Rows contain the
  input proxy, two prior works (A and B), and \cwm. The upper and lower cases extend
  to 298\,s and 299\,s, respectively, showing vehicle appearance and trajectory
  adherence over approximately five minutes.}
  \label{fig:long-horizon}
\end{figure}

Figures~\ref{fig:proxy-granularity}--\ref{fig:long-horizon} further demonstrate how the bidirectional creation checkpoint and the causal four-step checkpoint operate across representations, granularities, camera trajectories, and rollout horizons:
\clearpage
\newgeometry{includehead=false,top=1.35cm,bottom=2.5cm,
  left=2.5cm,right=2.5cm,columnsep=0.65cm,footskip=24pt}
\begingroup
  \setlength{\parskip}{0pt}
  \par\noindent
\begin{minipage}{\linewidth}
  \centering
  \captionsetup{font=small,skip=4pt,hypcap=false}
  \includegraphics[width=\linewidth]{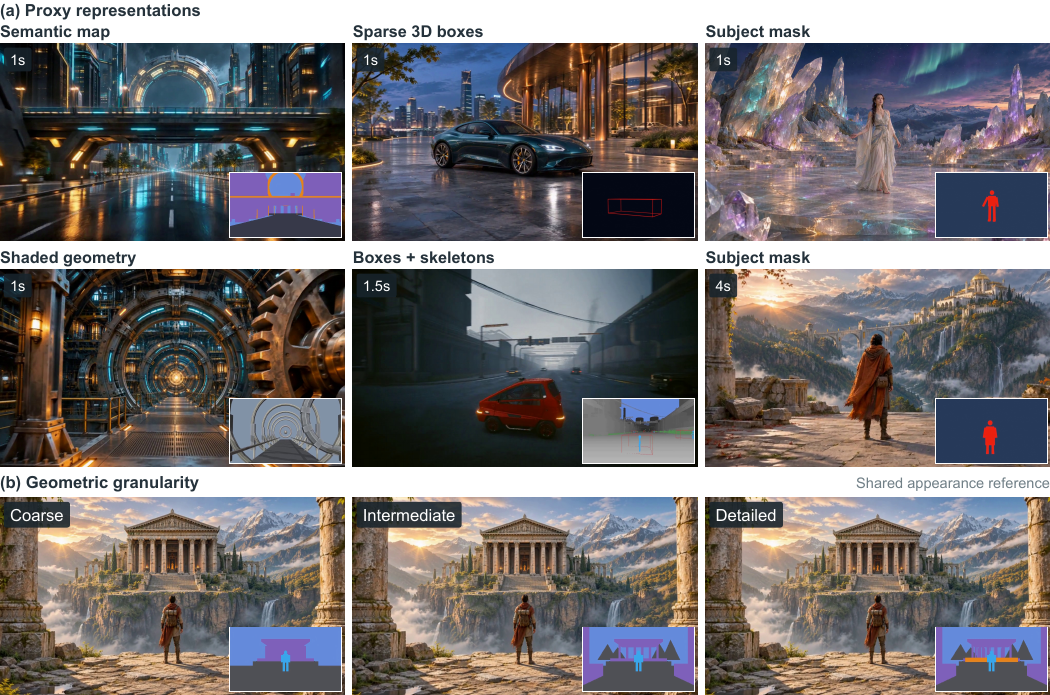}
  \captionof{figure}{\textbf{Control across proxy representations and geometric detail.}
  (a) Six examples spanning five proxy families, with synchronized input insets.
  (b) Coarse, intermediate, and detailed geometry under a shared appearance reference.}
  \label{fig:proxy-granularity}
\end{minipage}\par

  \vfill
  \par\noindent
\begin{minipage}{\linewidth}
  \centering
  \captionsetup{font=small,skip=4pt,hypcap=false}
  \includegraphics[width=\linewidth]{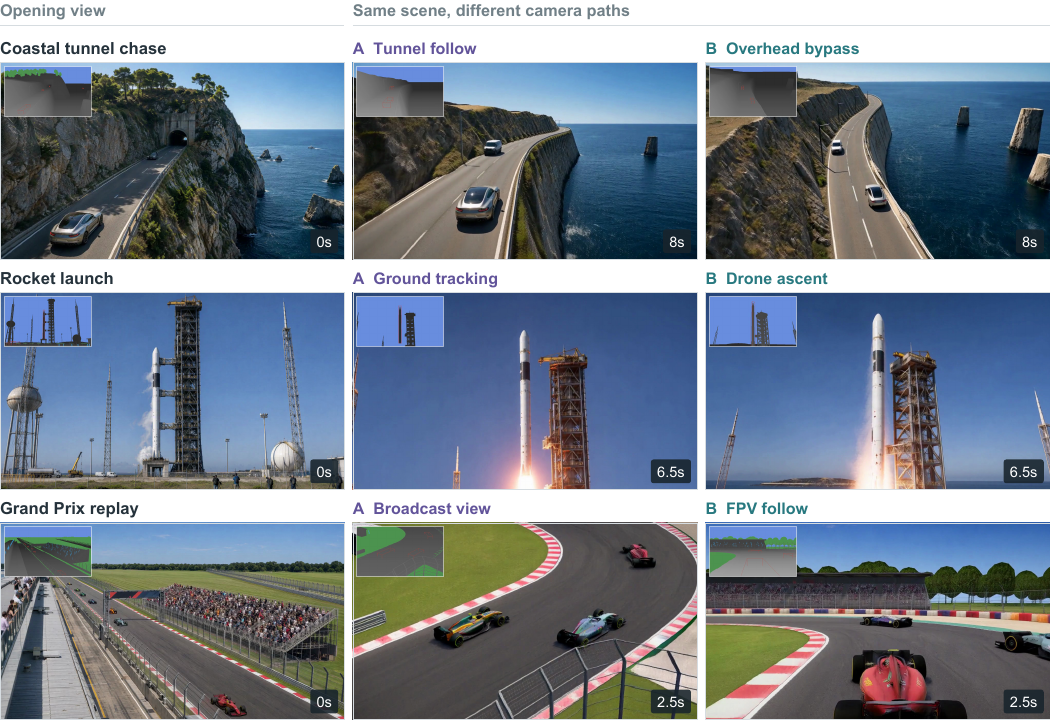}
  \captionof{figure}{\textbf{Two views of the same scene.}
  Each row shows an opening view followed by two synchronized outputs from
  different proxy-defined camera paths. Insets show the corresponding input proxies.}
  \label{fig:camera-control}
\end{minipage}\par

\par\endgroup
\clearpage
\restoregeometry

\begin{itemize}
    \item \textbf{Proxy families and geometric granularity (Figure~\ref{fig:proxy-granularity}).} Across five structural families (semantic maps, sparse 3D boxes, subject masks, shaded geometry, and boxes with skeletons), the creation model adheres to prescribed layouts and actor motion while synthesizing coherent lighting and surface detail (Figure~\ref{fig:proxy-granularity}a). Under a shared appearance reference, varying geometric granularity within one family (Figure~\ref{fig:proxy-granularity}b) transitions smoothly from global layout guidance under coarse geometry to exact surface anchoring under detailed geometry.
    \item \textbf{Camera-path control (Figure~\ref{fig:camera-control}).} Each row pairs two views of the same scene and event at a matched timestamp, using a shared appearance reference and different proxy-defined camera routes. Coastal driving contrasts tunnel following with an overhead bypass. The rocket launch pairs ground tracking with an ascending drone, and the racing replay pairs a broadcast view with FPV following. These examples show how the supplied camera trajectory changes the viewpoint while retaining the scene and subject appearance.
    \item \textbf{Long-horizon causal streaming (Figure~\ref{fig:long-horizon}).} Two racing sequences are compared at six matched timestamps each, extending to 298\,s and 299\,s. Driven incrementally by incoming proxies, the four-step causal model preserves vehicle appearance and follows the prescribed trajectories across successive streaming windows. These examples complement longer 30-minute racing sessions with consistent scene and vehicle appearance.
\end{itemize}

\FloatBarrier

\section{Conclusion}
\label{sec:conclusion}

We presented \cwm, a proxy-conditioned video rendering framework that unifies bidirectional video creation and four-step causal streaming under a shared multi-granularity interface. Built on a $\sim$200-hour multi-granularity video-proxy corpus, Progressive Forcing decouples causal adaptation and four-step DMD and GAN distillation on data-anchored histories from stateful continuation on self-generated rollouts, mitigating long-horizon drift while preserving structural fidelity and reference identity. Co-designing this renderer with a fused, KV-persistent, and codec-pipelined serving stack enables low-latency chunked execution, yielding consistent gains in visual quality, structural tracking, and reference preservation across streaming evaluations and sustaining coherent 30-minute racing rollouts toward infinite controllable generation.

Several directions remain open for scaling proxy-conditioned world rendering. First, although our static prefix sink and sliding KV window bound inference cost and sustain appearance over extended sessions, visual details outside the active window are eventually evicted once they leave the local context. Incorporating explicit long-term memory mechanisms directly into causal training and distillation, such as learnable memory compression, spatial-temporal context retrieval, or persistent 3D-aware state caches, would allow the renderer to recall previously visited environments and maintain multi-entity appearance consistency across revisits. Second, because current training uses a fixed proxy granularity within each rollout window, supporting dynamic mid-stream level-of-detail transitions and coupling \cwm with interactive state engines in closed loop remain promising avenues for future work.

\bibliographystyle{antgroup}
\bibliography{bibliography/references}

@misc{ge2026saltpp,
  title={{Salt++: Context-Aligned Post-Training for Few-Step Streaming Multimodal Generation}},
  author={Ge, Xingtong and Wang, Yutong and Zhu, Lunjie and Lin, Haitao and Lin, Fangyu and Huang, Yushi and Zhang, Xin and Zhang, Yi and Liu, Yu and Zhang, Jun},
  year={2026},
  eprint={2609.36995},
  archiveprefix={arXiv},
  url={https://arxiv.org/abs/2609.36995}
}

@misc{chen2024diffusionforcing,
  title={{Diffusion Forcing: Next-token Prediction Meets Full-Sequence Diffusion}},
  author={Chen, Boyuan and Marti Monso, Diego and Du, Yilun and Simchowitz, Max and Tedrake, Russ and Sitzmann, Vincent},
  year={2024},
  eprint={2407.01392},
  archiveprefix={arXiv},
  url={https://arxiv.org/abs/2407.01392}
}

@misc{yin2024dmd2,
  title={{Improved Distribution Matching Distillation for Fast Image Synthesis}},
  author={Yin, Tianwei and Gharbi, Micha{\"e}l and Park, Taesung and Zhang, Richard and Shechtman, Eli and Durand, Fredo and Freeman, William T.},
  year={2024},
  eprint={2405.14867},
  archiveprefix={arXiv},
  url={https://arxiv.org/abs/2405.14867}
}

@misc{zhu2026causalforcing,
  title={{Causal Forcing: Autoregressive Diffusion Distillation Done Right for High-Quality Real-Time Interactive Video Generation}},
  author={Zhu, Hongzhou and Zhao, Min and He, Guande and Su, Hang and Li, Chongxuan and Zhu, Jun},
  year={2026},
  eprint={2602.02214},
  archiveprefix={arXiv},
  url={https://arxiv.org/abs/2602.02214}
}

@misc{zhao2026causalforcingpp,
  title={{Causal Forcing++: Scalable Few-Step Autoregressive Diffusion Distillation for Real-Time Interactive Video Generation}},
  author={Zhao, Min and Zhu, Hongzhou and Zheng, Kaiwen and Zhou, Zihan and Yan, Bokai and Li, Xinyuan and Yang, Xiao and Li, Chongxuan and Zhu, Jun},
  year={2026},
  eprint={2605.15141},
  archiveprefix={arXiv},
  url={https://arxiv.org/abs/2605.15141}
}

@misc{zheng2026causalrcm,
  title={{Causal-rCM: A Unified Teacher-Forcing and Self-Forcing Open Recipe for Autoregressive Diffusion Distillation in Streaming Video Generation and Interactive World Models}},
  author={Zheng, Kaiwen and He, Guande and Zhao, Min and Zhang, Jintao and Chen, Huayu and Chen, Jianfei and Lin, Chen-Hsuan and Liu, Ming-Yu and Zhu, Jun and Ma, Qianli},
  year={2026},
  eprint={2606.25473},
  archiveprefix={arXiv},
  url={https://arxiv.org/abs/2606.25473}
}

@misc{minimaxh3,
  title={{MiniMax-H3}},
  author={{MiniMax-AI}},
  year={n.d.},
  howpublished={Official model repository},
  url={https://github.com/MiniMax-AI/MiniMax-H3},
  note={Accessed November 2026}
}

@misc{chen2025visual,
  title={{Visual Generation Without Guidance}},
  author={Chen, Huayu and Jiang, Kai and Zheng, Kaiwen and Chen, Jianfei and Su, Hang and Zhu, Jun},
  year={2025},
  eprint={2501.15420},
  archiveprefix={arXiv},
  url={https://arxiv.org/abs/2501.15420}
}

@inproceedings{yin2024dmd,
  title={{One-Step Diffusion with Distribution Matching Distillation}},
  author={Yin, Tianwei and Gharbi, Micha{\"e}l and Zhang, Richard and Shechtman, Eli and Durand, Fredo and Freeman, William T. and Park, Taesung},
  year={2024},
  booktitle={Proceedings of the IEEE/CVF Conference on Computer Vision and Pattern Recognition},
  url={https://arxiv.org/abs/2311.18828}
}

@misc{huang2025selfforcing,
  title={{Self Forcing: Bridging the Train-Test Gap in Autoregressive Video Diffusion}},
  author={Huang, Xun and Li, Zhengqi and He, Guande and Zhou, Mingyuan and Shechtman, Eli},
  year={2025},
  eprint={2506.08009},
  archiveprefix={arXiv},
  url={https://arxiv.org/abs/2506.08009}
}

@misc{yang2025longlive,
  title={{LongLive: Real-time Interactive Long Video Generation}},
  author={Yang, Shuai and Huang, Wei and Chu, Ruihang and Xiao, Yicheng and Zhao, Yuyang and Wang, Xianbang and Li, Muyang and Xie, Enze and Chen, Yingcong and Lu, Yao and Han, Song and Chen, Yukang},
  year={2025},
  eprint={2509.22622},
  archiveprefix={arXiv},
  url={https://arxiv.org/abs/2509.22622}
}

@misc{bruce2024genie,
  title={{Genie: Generative Interactive Environments}},
  author={Bruce, Jake and Dennis, Michael and Edwards, Ashley and Parker-Holder, Jack and Shi, Yuge and Hughes, Edward and Lai, Matthew and Mavalankar, Aditi and Steigerwald, Richie and Apps, Chris and Aytar, Yusuf and Bechtle, Sarah and Behbahani, Feryal and Chan, Stephanie and Heess, Nicolas and Gonzalez, Lucy and Osindero, Simon and Ozair, Sherjil and Reed, Scott and Zhang, Jingwei and Zolna, Konrad and Clune, Jeff and de Freitas, Nando and Singh, Satinder and Rockt{\"a}schel, Tim},
  year={2024},
  eprint={2402.15391},
  archiveprefix={arXiv},
  url={https://arxiv.org/abs/2402.15391}
}

@misc{valevski2024gamengen,
  title={{Diffusion Models Are Real-Time Game Engines}},
  author={Valevski, Dani and Leviathan, Yaniv and Arar, Moab and Fruchter, Shlomi},
  year={2024},
  eprint={2408.14837},
  archiveprefix={arXiv},
  url={https://arxiv.org/abs/2408.14837}
}

@misc{decart2024oasis,
  title={{Oasis: A Universe in a Transformer}},
  author={{Decart}},
  year={2024},
  howpublished={Project and technical release},
  url={https://oasis.decart.ai/}
}

@misc{wu2024ivideogpt,
  title={{iVideoGPT: Interactive VideoGPTs are Scalable World Models}},
  author={Wu, Jialong and Yin, Shaofeng and Feng, Ningya and He, Xu and Li, Dong and Hao, Jianye and Long, Mingsheng},
  year={2024},
  eprint={2405.15223},
  archiveprefix={arXiv},
  url={https://arxiv.org/abs/2405.15223}
}

@misc{wu2024pandora,
  title={{Pandora: Towards General World Model with Natural Language Actions and Video States}},
  author={Xiang, Jiannan and Liu, Guangyi and Gu, Yi and Gao, Qiyue and Ning, Yuting and Zha, Yuheng and Feng, Zeyu and Tao, Tianhua and Hao, Shibo and Shi, Yemin and Liu, Zhengzhong and Xing, Eric P. and Hu, Zhiting},
  year={2024},
  eprint={2406.09455},
  archiveprefix={arXiv},
  url={https://arxiv.org/abs/2406.09455}
}

@misc{yu2024gamegenx,
  title={{GameGen-X: Interactive Open-world Game Video Generation}},
  year={2024},
  eprint={2411.00769},
  archiveprefix={arXiv},
  url={https://arxiv.org/abs/2411.00769},
  author={Che, Haoxuan and He, Xuanhua and Liu, Quande and Jin, Cheng and Chen, Hao}
}

@misc{zhang2025matrixgame,
  title={{Matrix-Game: Interactive World Foundation Model}},
  author={Zhang, Yifan and Peng, Chunli and Wang, Boyang and Wang, Puyi and Zhu, Qingcheng and Kang, Fei and Jiang, Biao and Gao, Zedong and Li, Eric and Liu, Yang and Zhou, Yahui},
  year={2025},
  eprint={2506.18701},
  archiveprefix={arXiv},
  url={https://arxiv.org/abs/2506.18701}
}

@misc{mao2025yume,
  title={{Yume: An Interactive World Generation Model}},
  author={Mao, Xiaofeng and Lin, Shaoheng and Li, Zhen and Li, Chuanhao and Peng, Wenshuo and He, Tong and Pang, Jiangmiao and Chi, Mingmin and Qiao, Yu and Zhang, Kaipeng},
  year={2025},
  eprint={2507.17744},
  archiveprefix={arXiv},
  url={https://arxiv.org/abs/2507.17744}
}

@misc{sun2026worldplay,
  title={{WorldPlay: Towards Long-Term Geometric Consistency for Real-Time Interactive World Modeling}},
  author={Sun, Wenqiang and Zhang, Haiyu and Wang, Haoyuan and Wu, Junta and Wang, Zehan and Wang, Zhenwei and Wang, Yunhong and Zhang, Jun and Wang, Tengfei and Guo, Chunchao},
  year={2025},
  eprint={2512.14614},
  archiveprefix={arXiv},
  url={https://arxiv.org/abs/2512.14614}
}

@misc{robbyant2026lingbot,
  title={{Advancing Open-source World Models}},
  author={{Robbyant Team} and Gao, Zelin and Wang, Qiuyu and Zeng, Yanhong and Zhu, Jiapeng and Cheng, Ka Leong and Li, Yixuan and Wang, Hanlin and Xu, Yinghao and Ma, Shuailei and Chen, Yihang and Liu, Jie and Cheng, Yansong and Yao, Yao and Zhu, Jiayi and Meng, Yihao and Zheng, Kecheng and Bai, Qingyan and Chen, Jingye and Shen, Zehong and Yu, Yue and Zhu, Xing and Shen, Yujun and Ouyang, Hao},
  year={2026},
  eprint={2601.20540},
  archiveprefix={arXiv},
  url={https://arxiv.org/abs/2601.20540}
}

@misc{ma2026lingbotvideo,
  title={{Scaling Mixture-of-Experts Video Pretraining for Embodied Intelligence}},
  author={Ma, Shuailei and Liao, Jiaqi and Wang, Xinyang and Wang, Jingjing and Feng, Chaoran and Hu, Zijing and Bao, Chong and Xi, Zichen and Gan, Yuqi and Wang, Weisen and Zeng, Yanhong and Zhao, Qin and Shi, Zifan and Wu, Wei and Ouyang, Hao and Wang, Qiuyu and Zhang, Shangzhan and Shao, Jiahao and Sun, Yipengjing and Hu, Liangxiao and Pan, Lunke and Xue, Nan and Zheng, Kecheng and Xu, Yinghao and Zhu, Xing and Shen, Yujun and Cheng, Ka Leong},
  year={2026},
  eprint={2607.07675},
  archiveprefix={arXiv},
  url={https://arxiv.org/abs/2607.07675}
}

@misc{zhang2026matrixgame4,
  title={{Matrix-Game 4.0: Harnessing Video Models for Real-Time Playable Game Worlds}},
  author={Zhang, Jihai and Qian, Runjia and Kong, Kaiqin and Jiang, Biao and Wang, Puyi and Chen, Siru and Li, Tianle and Chen, Guanjie and Wu, Junyi and Chen, Chubin and Li, Yuanhang and Huang, Kaichen and Zhang, Haobo and Li, Yangguang and Yu, Wei and Chen, Guibin and Cheng, Yu and Zhou, Yahui},
  year={2026},
  eprint={2610.27999},
  archiveprefix={arXiv},
  url={https://arxiv.org/abs/2610.27999}
}

@misc{gao2026infiniteworlds,
  title={{Infinite Worlds with Versatile Interactions}},
  author={Gao, Zelin and Wang, Qiuyu and Zhu, Jiapeng and Chen, Jingye and Liu, Zichen and Bai, Qingyan and Wang, Jiahao and Yuan, Yufeng and Wang, Hanlin and Lu, Yichong and Cheng, Ka Leong and Zhang, Haojie and Gao, Jian and Feng, Tianrui and Liu, Yuzheng and Yao, Yao and Xu, Yinghao and Zhu, Xing and Shen, Yujun and Ouyang, Hao},
  year={2026},
  eprint={2607.07534},
  archiveprefix={arXiv},
  url={https://arxiv.org/abs/2607.07534}
}

@misc{xu2026proxy2world,
  title={{Proxy2World: Learning to Generate Worlds From Lightweight Proxies without Seeing Them}},
  author={Xu, Hongli and Yan, Weilong and Wang, Anbang and Zou, Chunyu and Hong, Siyu and Huang, Jingwei},
  year={2026},
  eprint={2609.35023},
  archiveprefix={arXiv},
  url={https://arxiv.org/abs/2609.35023}
}

@misc{liang2026codevideomodel,
  title={{Code Video Model}},
  author={Liang, Zhiyang and Zhang, Jiawei and Qian, Guocheng and Chen, Dongdong and Wan, Ziyu},
  year={2026},
  howpublished={Technical report and project page},
  url={https://code-video-model.github.io/}
}

@misc{chen2026cwm,
  title={{Code World Model: Coding Agent as World Brain}},
  author={Chen, Yiwen and Lin, Guosheng and Zhang, Chi},
  year={2026},
  eprint={2608.25927},
  archiveprefix={arXiv},
  url={https://arxiv.org/abs/2608.25927}
}

@misc{huang2026pwm,
  title={{Programmable World Model}},
  author={Huang, Zheng-Hui and Lin, Guixu and Lin, Jiacheng and Huang, Yi-Chuan and Yu, Ruihan and Niu, Muyao and Yang, Siqi and Liu, Yu-Lun and Chuang, Yung-Yu and Zhang, Kaipeng and Wang, Zhixiang},
  year={2026},
  eprint={2609.10540},
  archiveprefix={arXiv},
  url={https://arxiv.org/abs/2609.10540}
}

@misc{lu2026streamrenderh3,
  title={{StreamRender-H3: Interactive Coding Worlds with Real-Time Streaming Diffusion Render}},
  author={Lu, Yanzuo and Ye, Tian and Li, Yitong and Xue, Shuchen and Liu, Haozhe and Han, Song and Xie, Enze},
  year={2026},
  howpublished={NVIDIA Research project page},
  url={https://nvlabs.github.io/Sana/Sol-Engine/StreamRender-H3/}
}

@misc{liao2026streamingcwm,
  title={{Streaming Code World Model}},
  author={Liao, Tingting and Liu, Guangyi},
  year={2026},
  howpublished={GitHub repository and model release},
  url={https://github.com/TingtingLiao/streaming-code-world-model}
}

@misc{zhan2026magpie,
  title={{Magpie: Real-Time World Renderer for Interactive Games}},
  author={Zhan, Xiaoyu and Wang, Xinyu and Zhang, Xiaohong and Zhu, Huanjie and Sun, Tengjiao and Fang, Pengcheng and Yu, Jiaxing and Guo, Yanwen and Fu, Dongjie},
  year={2026},
  eprint={2608.27168},
  archiveprefix={arXiv},
  url={https://arxiv.org/abs/2608.27168}
}

@misc{zhou2025omniworld,
  title={{OmniWorld: A Multi-Domain and Multi-Modal Dataset for 4D World Modeling}},
  author={Zhou, Yang and Wang, Yifan and Zhou, Jianjun and Chang, Wenzheng and Guo, Haoyu and Li, Zizun and Ma, Kaijing and Li, Xinyue and Wang, Yating and Zhu, Haoyi and Liu, Mingyu and Liu, Dingning and Yang, Jiange and Fu, Zhoujie and Chen, Junyi and Shen, Chunhua and Pang, Jiangmiao and Zhang, Kaipeng and He, Tong},
  year={2025},
  eprint={2509.12201},
  archiveprefix={arXiv},
  url={https://arxiv.org/abs/2509.12201}
}

@inproceedings{dosovitskiy2017carla,
  title={{CARLA: An Open Urban Driving Simulator}},
  author={Dosovitskiy, Alexey and Ros, German and Codevilla, Felipe and Lopez, Antonio and Koltun, Vladlen},
  booktitle={Proceedings of the 1st Annual Conference on Robot Learning},
  series={Proceedings of Machine Learning Research},
  volume={78},
  pages={1--16},
  year={2017},
  url={https://proceedings.mlr.press/v78/dosovitskiy17a.html}
}

@misc{chi2026agentgarten,
  title={{AgentGarten: Code Worlds for Evolving Agents}},
  author={Chi, Jiawei and Miao, Shangchen and Shi, Zhiyuan and Wu, Kailu and Wang, Hanyang and Chen, Weiliang and Dai, Qiyu and Li, Chengyang and Ren, Jinshan and Gao, Jun and Long, Mingsheng and Duan, Yueqi and Lyu, Jiangran and Wu, Jialong and Liu, Fangfu},
  year={2026},
  eprint={2610.12374},
  archiveprefix={arXiv},
  url={https://arxiv.org/abs/2610.12374}
}

@misc{luo2026mundo,
  title={{Mundo: Customized World Model}},
  author={Luo, Yihang and Xu, Zunnan and Zhou, Shangchen and Lan, Yushi and Liao, Kang and Gong, Zerui and Wu, Xiaoming and Tao, Qingyi and Wu, Zhonghua and Jin, Bixuan and Wang, Fei and Hou, Zhi and Wang, Xiaogang and Pan, Xingang and Loy, Chen Change},
  year={2026},
  howpublished={Technical report and project page},
  url={https://mundo-world-model.github.io/}
}

@misc{yang2026oneira,
  title={{Oneira: From Open-Ended Generation to Open-World Interaction in Video World Models}},
  author={Yang, Xindi and Li, Baolu and Lee, Liam and Yin, Zhenfei and Zhang, Songxin and Song, Zhuoyang and Jia, Xu and Cai, Jianfei and Wong, Tien-Tsin and Jing, Bingyi and Yang, Mengyue},
  year={2026},
  eprint={2610.01614},
  archiveprefix={arXiv},
  url={https://arxiv.org/abs/2610.01614}
}

@misc{wu2026boxworld,
  title={{BoxWorld: Let World Engines Control and Video Generators Render}},
  author={Wu, Zike and Hatem, Ahmed and Trivedi, Chintan and Lira, Wallace Michel Pinto and Qin, Xuebin and Wang, Lele and Liao, Renjie},
  year={2026},
  eprint={2610.36371},
  archiveprefix={arXiv},
  url={https://arxiv.org/abs/2610.36371}
}

@misc{li2026fightworld,
  title={{Fight World Model: Customizable Combat through Visual Interaction Proxies}},
  author={Li, Jiahua and John, Zixu and Zhong, Tom and Wu, Fuping and Xu, Tianhao and Zhang, Ruicheng and Zhou, Jun and Xu, Zunnan},
  year={2026},
  eprint={2610.40607},
  archiveprefix={arXiv},
  url={https://arxiv.org/abs/2610.40607}
}

@misc{fang2026solarrenderer,
  title={{SolarRenderer: Few-Step Generative Reference Rendering}},
  author={Fang, Guian and Qian, Shengju and Lu, Dongyue and Huang, Junchao and Jiang, Lutao and Wang, Xin and Song, Yiren and Jiang, Li and Shou, Mike Zheng},
  year={2026},
  eprint={2610.28630},
  archiveprefix={arXiv},
  url={https://arxiv.org/abs/2610.28630}
}

@misc{liu2026anygamerenderer,
  title={{AnyGameRenderer: Entity-Consistent Generative Rendering Across Games and an Entity-Level Benchmark}},
  author={Liu, Ruicong and Ouyang, Liangyang and Lan, Fengbo and Wang, Zhixiang and Zhang, Kaipeng},
  year={2026},
  eprint={2610.31466},
  archiveprefix={arXiv},
  url={https://arxiv.org/abs/2610.31466}
}

@misc{huang2026edenlive,
  title={{EdenLive: Real-Time Generative Rendering from Any Visual Reference}},
  author={Huang, Zehuan and Xu, Tianxing and Peng, Chunli and Zhang, Cheng and Zheng, Zeyi and Song, Yifei and Gu, Zekai and Yu, Zhipeng and Wang, Peng and Cao, Yanpei},
  year={2026},
  howpublished={Technical report and open-source project},
  url={https://github.com/VAST-AI-Research/EdenLive}
}

@misc{meng2026marionette,
  title={{Marionette: Predicting World States, Rendering Geometry, Painting Appearance}},
  author={Meng, Zian and Li, Zhen and Li, Chuanhao and Li, Qiang and Zhang, Kaipeng},
  year={2026},
  eprint={2608.14530},
  archiveprefix={arXiv},
  url={https://arxiv.org/abs/2608.14530}
}

@misc{wu2026wizard,
  title={{Wizard: Manipulate Any World in Real Time via Agentic Control}},
  author={Wu, Yuzhe and Lin, Zijun and Jin, Yeying},
  year={2026},
  eprint={2610.40339},
  archiveprefix={arXiv},
  url={https://arxiv.org/abs/2610.40339}
}

@misc{huang2026generativeworldrenderer,
  title={{Generative World Renderer}},
  author={Huang, Zheng-Hui and Wang, Zhixiang and Tan, Jiaming and Yu, Ruihan and Zhang, Yidan and Zheng, Bo and Liu, Yu-Lun and Chuang, Yung-Yu and Zhang, Kaipeng},
  year={2026},
  eprint={2604.02329},
  archiveprefix={arXiv},
  url={https://arxiv.org/abs/2604.02329}
}

@misc{chen2026dar,
  title={{Video Models as Native 4D Renderers: World-Grounded Conditioning from Animated Mesh}},
  author={Chen, Junhao and Chen, Mingjin and Zhang, Henghaofan and Chen, Minglin and Fan, Liaoyuan and Zhang, Boran and Zhang, Saining and Sun, Mingze and Zhao, Hao and Huang, Ruqi and Li, Zhihao and Wang, Yufei},
  year={2026},
  eprint={2608.00094},
  archiveprefix={arXiv},
  url={https://arxiv.org/abs/2608.00094}
}

@misc{he2025matrixgame2,
  title={{Matrix-game 2.0: An open-source, real-time, and streaming interactive world model}},
  year={2025},
  eprint={2508.13009},
  archiveprefix={arXiv},
  url={https://arxiv.org/abs/2508.13009},
  author={He, Xianglong and Peng, Chunli and Liu, Zexiang and Wang, Boyang and Zhang, Yifan and Cui, Qi and Kang, Fei and Jiang, Biao and An, Mengyin and Ren, Yangyang and Xu, Baixin and Guo, Hao-Xiang and Gong, Kaixiong and Wu, Size and Li, Wei and Song, Xuchen and Liu, Yang and Li, Yangguang and Zhou, Yahui}
}

@misc{wang2026matrixgame3,
  title={{Matrix-Game 3.0: Real-Time and Streaming Interactive World Model with Long-Horizon Memory}},
  year={2026},
  eprint={2604.08995},
  archiveprefix={arXiv},
  url={https://arxiv.org/abs/2604.08995},
  author={Wang, Zile and Liu, Zexiang and Li, Jiaxing and Huang, Kaichen and Xu, Baixin and Kang, Fei and An, Mengyin and Wang, Peiyu and Jiang, Biao and Wei, Yichen and Xietian, Yidan and Pei, Jiangbo and Hu, Liang and Jiang, Boyi and Xue, Hua and Wang, Zidong and Sun, Haofeng and Li, Wei and Ouyang, Wanli and He, Xianglong and Liu, Yang and Li, Yangguang and Zhou, Yahui}
}

@misc{zhu2026sanawm,
  title={{SANA-WM: Efficient Minute-Scale World Modeling with Hybrid Linear Diffusion Transformer}},
  author={Zhu, Haoyi and Liu, Haozhe and Zhao, Yuyang and Ye, Tian and Chen, Junsong and Yu, Jincheng and He, Tong and Han, Song and Xie, Enze},
  year={2026},
  eprint={2605.15178},
  archiveprefix={arXiv},
  url={https://arxiv.org/abs/2605.15178}
}

@misc{zhao2026minwm,
  title={{minWM: A Full-Stack Open-Source Framework for Real-Time Interactive Video World Models}},
  author={Zhao, Min and Zhu, Hongzhou and Yan, Bokai and Zhou, Zihan and Chen, Yimin and Wang, Honglie and Sun, Wenqiang and Fang, Zhengwei and Xun, Zizheng and Li, Zihao and Zheng, Kaiwen and He, Guande and Yang, Xiao and Li, Chongxuan and Bao, Fan and Zhu, Jun},
  year={2026},
  eprint={2605.30263},
  archiveprefix={arXiv},
  url={https://arxiv.org/abs/2605.30263}
}

@misc{qian2026matrixgame35,
  title={{Matrix-Game 3.5: Enhancing Real-Time Streaming Interactive World Models with Patch Memory}},
  author={Qian, Runjia and Wang, Zile and Zhang, Jihai and Zou, Kai and Yu, Wei and Li, Jiaxing and Liu, Zexiang and Li, Yaokun and Kang, Fei and Huang, Kaichen and An, Mengyin and Zhang, Haobo and Jiang, Biao and Wang, Jiahua and Sun, Haofeng and Liu, Yang and Li, Yangguang},
  year={2026},
  eprint={2608.29910},
  archiveprefix={arXiv},
  url={https://arxiv.org/abs/2608.29910}
}

@misc{yu2026worldcrafter,
  title={{WorldCrafter: Consistent Video World Model with Implicit 3D-aware Memory}},
  author={Yu, Wangbo and Liu, Kunhao and Hu, Wenbo and Yuan, Shenghai and Feng, Chaoran and Zhou, Haiyang and Huang, Yukun and Wang, Yiran and Zhao, Wang and Luo, Yingmin and Shan, Ying},
  year={2026},
  eprint={2609.24984},
  archiveprefix={arXiv},
  url={https://arxiv.org/abs/2609.24984}
}

@misc{lin2026alayaflash,
  title={{Generative World Renderer at the Speed of Play}},
  author={Lin, Guixu and Huang, Zheng-Hui and Yang, Siqi and Yang, Ming-Hsuan and Zhang, Kaipeng and Wang, Zhixiang},
  year={2026},
  eprint={2607.18703},
  archiveprefix={arXiv},
  url={https://arxiv.org/abs/2607.18703}
}

@article{cabon2020virtualkitti2,
  title = {Virtual {KITTI} 2},
  author = {Cabon, Yohann and Murray, Naila and Humenberger, Martin},
  journal = {arXiv preprint arXiv:2001.10773},
  year = {2020},
  url = {https://arxiv.org/abs/2001.10773}
}

@inproceedings{zhang2023controlnet,
  title={{Adding Conditional Control to Text-to-Image Diffusion Models}},
  author={Zhang, Lvmin and Rao, Anyi and Agrawala, Maneesh},
  booktitle={Proceedings of the IEEE/CVF International Conference on Computer Vision},
  year={2023},
  url={https://arxiv.org/abs/2302.05543}
}

@inproceedings{liang2025diffusionrenderer,
  title={{DiffusionRenderer: Neural Inverse and Forward Rendering with Video Diffusion Models}},
  author={Liang, Ruofan and Gojcic, Zan and Ling, Huan and Munkberg, Jacob and Hasselgren, Jon and Lin, Zhi-Hao and Gao, Jun and Keller, Alexander and Vijaykumar, Nandita and Fidler, Sanja and Wang, Zian},
  booktitle={Proceedings of the IEEE/CVF Conference on Computer Vision and Pattern Recognition},
  year={2025},
  url={https://arxiv.org/abs/2501.18590}
}

@misc{jiang2025vace,
  title={{VACE: All-in-One Video Creation and Editing}},
  author={Jiang, Zeyinzi and Han, Zhen and Mao, Chaojie and Zhang, Jingfeng and Pan, Yulin and Liu, Yu},
  year={2025},
  eprint={2503.07598},
  archiveprefix={arXiv},
  url={https://arxiv.org/abs/2503.07598}
}

@inproceedings{he2025cameractrl,
  title={{CameraCtrl: Enabling Camera Control for Text-to-Video Generation}},
  author={He, Hao and Xu, Yinghao and Guo, Yuwei and Wetzstein, Gordon and Dai, Bo and Li, Hongsheng and Yang, Ceyuan},
  booktitle={International Conference on Learning Representations},
  year={2025},
  url={https://arxiv.org/abs/2404.02101}
}

@inproceedings{yin2025causvid,
  title={{From Slow Bidirectional to Fast Autoregressive Video Diffusion Models}},
  author={Yin, Tianwei and Zhang, Qiang and Zhang, Richard and Freeman, William T. and Durand, Fredo and Shechtman, Eli and Huang, Xun},
  booktitle={Proceedings of the IEEE/CVF Conference on Computer Vision and Pattern Recognition},
  year={2025},
  url={https://arxiv.org/abs/2412.07772}
}

@inproceedings{kim2024fifodiffusion,
  title={{FIFO-Diffusion: Generating Infinite Videos from Text without Training}},
  author={Kim, Jihwan and Kang, Junoh and Choi, Jinyoung and Han, Bohyung},
  booktitle={Advances in Neural Information Processing Systems},
  year={2024},
  url={https://arxiv.org/abs/2405.11473}
}

@inproceedings{song2023consistencymodels,
  title={{Consistency Models}},
  author={Song, Yang and Dhariwal, Prafulla and Chen, Mark and Sutskever, Ilya},
  booktitle={Proceedings of the 40th International Conference on Machine Learning},
  year={2023},
  url={https://arxiv.org/abs/2303.01469}
}

@misc{sauer2024ladd,
  title={{Fast High-Resolution Image Synthesis with Latent Adversarial Diffusion Distillation}},
  author={Sauer, Axel and Boesel, Frederic and Dockhorn, Tim and Blattmann, Andreas and Esser, Patrick and Rombach, Robin},
  year={2024},
  eprint={2403.12015},
  archiveprefix={arXiv},
  url={https://arxiv.org/abs/2403.12015}
}

@inproceedings{yang2024depthanythingv2,
  title={{Depth Anything V2}},
  author={Yang, Lihe and Kang, Bingyi and Huang, Zilong and Zhao, Zhen and Xu, Xiaogang and Feng, Jiashi and Zhao, Hengshuang},
  booktitle={Advances in Neural Information Processing Systems},
  year={2024},
  url={https://arxiv.org/abs/2406.09414}
}

@inproceedings{ravi2024sam2,
  title={{SAM 2: Segment Anything in Images and Videos}},
  author={Ravi, Nikhila and Gabeur, Valentin and Hu, Yuan-Ting and Hu, Ronghang and Ryali, Chaitanya and Ma, Tengyu and Khedr, Haitham and R{\"a}dle, Roman and Rolland, Chloe and Gustafson, Laura and Mintun, Eric and Pan, Junting and Alwala, Kalyan Vasudev and Carion, Nicolas and Wu, Chao-Yuan and Girshick, Ross and Doll{\'a}r, Piotr and Feichtenhofer, Christoph},
  booktitle={International Conference on Learning Representations},
  year={2025},
  url={https://arxiv.org/abs/2408.00714}
}

@inproceedings{peebles2023dit,
  title={{Scalable Diffusion Models with Transformers}},
  author={Peebles, William and Xie, Saining},
  booktitle={Proceedings of the IEEE/CVF International Conference on Computer Vision},
  year={2023},
  url={https://arxiv.org/abs/2212.09748}
}

@article{su2024roformer,
  title={{RoFormer: Enhanced Transformer with Rotary Position Embedding}},
  author={Su, Jianlin and Ahmed, Murtadha and Lu, Yu and Pan, Shengfeng and Bo, Wen and Liu, Yunfeng},
  journal={Neurocomputing},
  volume={568},
  pages={127063},
  year={2024},
  url={https://arxiv.org/abs/2104.09864}
}

@inproceedings{lipman2023flowmatching,
  title={{Flow Matching for Generative Modeling}},
  author={Lipman, Yaron and Chen, Ricky T. Q. and Ben-Hamu, Heli and Nickel, Maximilian and Le, Matt},
  booktitle={International Conference on Learning Representations},
  year={2023},
  url={https://arxiv.org/abs/2210.02747}
}

@inproceedings{liu2023rectifiedflow,
  title={{Flow Straight and Fast: Learning to Generate and Transfer Data with Rectified Flow}},
  author={Liu, Xingchao and Gong, Chengyue and Liu, Qiang},
  booktitle={International Conference on Learning Representations},
  year={2023},
  url={https://arxiv.org/abs/2209.03003}
}

@inproceedings{jolicoeurmartineau2019relativistic,
  title={{The relativistic discriminator: a key element missing from standard GAN}},
  author={Jolicoeur-Martineau, Alexia},
  booktitle={International Conference on Learning Representations},
  year={2019},
  url={https://arxiv.org/abs/1807.00734}
}

@inproceedings{xiao2024streamingllm,
  title={{Efficient Streaming Language Models with Attention Sinks}},
  author={Xiao, Guangxuan and Tian, Yuandong and Chen, Beidi and Han, Song and Lewis, Mike},
  booktitle={International Conference on Learning Representations},
  year={2024},
  url={https://arxiv.org/abs/2309.17453}
}

@misc{ho2022cfg,
  title={{Classifier-Free Diffusion Guidance}},
  author={Ho, Jonathan and Salimans, Tim},
  year={2022},
  eprint={2207.12598},
  archiveprefix={arXiv},
  url={https://arxiv.org/abs/2207.12598}
}

@inproceedings{loshchilov2019adamw,
  title={{Decoupled Weight Decay Regularization}},
  author={Loshchilov, Ilya and Hutter, Frank},
  booktitle={International Conference on Learning Representations},
  year={2019},
  url={https://arxiv.org/abs/1711.05101}
}

@article{bradley1952rank,
  title={{Rank Analysis of Incomplete Block Designs: I. The Method of Paired Comparisons}},
  author={Bradley, Ralph Allan and Terry, Milton E.},
  journal={Biometrika},
  volume={39},
  number={3/4},
  pages={324--345},
  year={1952}
}

@inproceedings{luo2023diffinstruct,
  title={{Diff-Instruct: A Universal Approach for Transferring Knowledge From Pre-trained Diffusion Models}},
  author={Luo, Weijian and Hu, Tianyang and Zhang, Shifeng and Sun, Jiacheng and Li, Zhenguo and Zhang, Zhihua},
  booktitle={Advances in Neural Information Processing Systems},
  year={2023},
  url={https://arxiv.org/abs/2305.18455}
}

@misc{sauer2023add,
  title={{Adversarial Diffusion Distillation}},
  author={Sauer, Axel and Lorenz, Dominik and Blattmann, Andreas and Rombach, Robin},
  year={2023},
  eprint={2311.17042},
  archiveprefix={arXiv},
  url={https://arxiv.org/abs/2311.17042}
}

@inproceedings{zhou2024sid,
  title={{Score Identity Distillation: Exponentially Fast Distillation of Pretrained Diffusion Models for One-Step Generation}},
  author={Zhou, Mingyuan and Zheng, Huangjie and Wang, Zhendong and Yin, Mingzhang and Huang, Hai},
  booktitle={International Conference on Machine Learning},
  year={2024},
  url={https://arxiv.org/abs/2404.04057}
}

@misc{luo2025tdm,
  title={{Learning Few-Step Diffusion Models by Trajectory Distribution Matching}},
  author={Luo, Yihong and Hu, Tianyang and Sun, Jiacheng and Cai, Yujun and Tang, Jing},
  year={2025},
  eprint={2503.06674},
  archiveprefix={arXiv},
  url={https://arxiv.org/abs/2503.06674}
}

@misc{luo2026vatdm,
  title={{VA-TDM: A Unified View for Improving Few-Step Diffusion Models by Velocity Adversarial Distillation}},
  author={Luo, Yihong and Pang, Tianyu and Hu, Tianyang},
  year={2026},
  eprint={2611.06796},
  archiveprefix={arXiv},
  url={https://arxiv.org/abs/2611.06796}
}

@misc{zhao2026edmd,
  title={{eDMD: Endpoint Distribution Matching for Few-Step Diffusion Distillation}},
  author={Zhao, Tong and Zeng, Xianfang and Han, Yucheng and Wang, Ruoyu and Lei, Mingkun and Yu, Gang and Zhang, Chi},
  year={2026},
  eprint={2611.17194},
  archiveprefix={arXiv},
  url={https://arxiv.org/abs/2611.17194}
}
\end{document}